\documentclass{article}
\usepackage{wrapfig}
\usepackage{amsmath}
\usepackage{algorithm}
\usepackage{algpseudocode}  
\usepackage{soul}    

\usepackage{xcolor}  
\definecolor{darkgreen}{RGB}{0,128,0}

\usepackage{url}
\definecolor{citeblue}{rgb}{0.21,0.49,0.74}
\usepackage[pagebackref=false,breaklinks,colorlinks,citecolor=citeblue,bookmarks=false]{hyperref}
\PassOptionsToPackage{numbers, compress}{natbib}
\usepackage[preprint]{neurips_2026}

\usepackage[utf8]{inputenc} 
\usepackage[T1]{fontenc}    
\usepackage{hyperref}       
\usepackage{url}            
\usepackage{booktabs}       
\usepackage{amsfonts}       
\usepackage{nicefrac}       
\usepackage{microtype}      
\usepackage{xcolor}         
\usepackage{eso-pic}       
\usepackage{graphicx}      
\usepackage{calc}          
\usepackage{colortbl}
\usepackage{xspace}
\newcolumntype{C}{>{\centering\arraybackslash}p{0.04\textwidth}}

\usepackage[listings]{tcolorbox}
\tcbuselibrary{breakable}
\usepackage{enumitem}
\usepackage{pifont}
\usepackage{fontawesome}  
\newcommand{\lightgraybg}{\rowcolor{gray!10}}
\usepackage{booktabs}
\usepackage{multirow} 
\newcommand{\graybg}{\rowcolor{gray!20}}

\newcommand{\colorEvidence}[1]{\textcolor[rgb]{0.13,0.67,0.8}{\textbf{#1}}}
\newcommand{\token}[1]{\texttt{\small\textless#1\textgreater}}
\newcommand{\upred}[1]{\textcolor{purple}{\tiny \,$\uparrow$#1}}
\newcommand{\downblue}[1]{\textcolor{blue}{\tiny \,$\downarrow$#1}}

\newtcolorbox{templatebox}[1]{
  enhanced,
  unbreakable,
  colback=white,
  colframe=black!65,
  colbacktitle=black!80,
  coltitle=white,
  boxrule=0.9pt,
  arc=2pt,
  left=6pt,
  right=6pt,
  top=6pt,
  bottom=6pt,
  title={#1},
  fonttitle=\bfseries,
  sharp corners,
  boxed title style={sharp corners, boxrule=0pt}
}
\usepackage{tcolorbox}
\tcbuselibrary{skins,breakable}
\definecolor{boxbackground}{HTML}{F0F7FF}
\definecolor{boxborder}{HTML}{D0D9E5}
\definecolor{accentblue}{HTML}{4A86E8}

\newtcolorbox{promptbox}[2][]{%
    enhanced,
    breakable,
    boxsep=5pt,
    left=9pt,
    right=7pt,
    top=5pt,
    bottom=5pt,
    colback=boxbackground,
    colframe=boxborder,
    boxrule=0.5pt,
    arc=4pt,
    frame hidden,
    borderline west={3pt}{0pt}{accentblue},
    shadow={0.5pt}{0.5pt}{1.5pt}{black!10},
    fontupper=\normalsize,
    title=#2,
    colbacktitle=accentblue,
    coltitle=white,
    fonttitle={\fontsize{9}{11}\selectfont\bfseries},
    attach boxed title to top left={yshift=-2.5mm, xshift=3.2mm},
    boxed title style={
        enhanced,
        left=3pt,
        right=3pt,
        top=1pt,
        bottom=1pt,
        boxsep=2pt,
        arc=3pt,
        boxrule=0pt,
        colback=accentblue,
    },
    #1
}

\usepackage{amsthm}

\usepackage{amsmath}
\usepackage{amssymb}

\usepackage{thmtools}
\theoremstyle{plain}

\theoremstyle{definition}

\theoremstyle{remark}

\AddToShipoutPictureBG{%
  \ifnum\value{page}=1
    \AtPageUpperLeft{%
      \put(\LenToUnit{3.8cm},\LenToUnit{-2.7cm}){%
        \includegraphics[height=0.8cm]{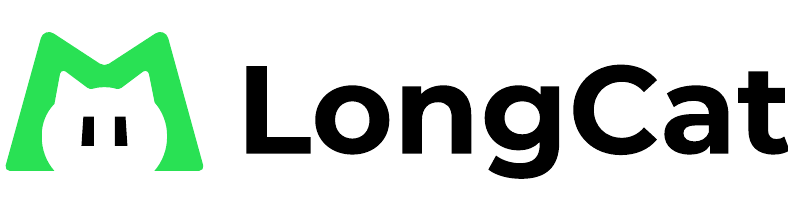}%
      }%
    }%
  \fi
}
\title{MAP: A Map-then-Act Paradigm for Long-Horizon Interactive Agent Reasoning}

\author{
  \textbf{Yuxin Liu$^{1,2\star}$},
  \textbf{Ziang Ye$^{1,2}$},
  \textbf{Yueqing Sun$^{2}$},
  \textbf{Mingye Zhu$^{1}$},
  \textbf{Jinwei Xiao$^{2,3}$},
  \\
  \textbf{Zhuowen Han$^{2,4}$},
  \textbf{Qi Gu$^{2,\dagger}$},
  \textbf{Xunliang Cai$^{2}$},
  \textbf{Lei Zhang$^{1,\dagger}$} \\
  \vspace{-2mm} \\
  $^{1}$University of Science and Technology of China 
  $^{2}$Meituan \\
  $^{3}$Institution of Automation, Chinese Academy of Sciences
  $^{4}$Tianjin University\\
  \texttt{\small liuyuxin1010@mail.ustc.edu.cn}, \texttt{\small  leizh23@ustc.edu.cn}, \texttt{\small guqi03@meituan.com}
}

\begin{document}

\maketitle
\vspace{-16pt}
\begin{abstract}

Current interactive LLM agents rely on goal-conditioned stepwise planning, where environmental understanding is acquired reactively during execution rather than established beforehand. This temporal inversion leads to Delayed Environmental Perception: agents must infer environmental constraints through trial-and-error, resulting in an Epistemic Bottleneck that traps them in inefficient failure cycles.
Inspired by human affordance perception and cognitive map theory, we propose the \textbf{M}ap-then-\textbf{A}ct \textbf{P}aradigm (MAP), a plug-and-play framework that shifts environment understanding before execution. MAP consists of three stages: (1) Global Exploration, acquiring environment-general priors; (2) Task-Specific Mapping, constructing a structured cognitive map; and (3) Knowledge-Augmented Execution, solving tasks grounded on the map.
Experiments show consistent gains across benchmarks and LLMs. On ARC-AGI-3, MAP enables frontier models to surpass near-zero baseline performance in 22 of 25 game environments. We further introduce MAP-2K, a dataset of map-then-act trajectories, and show that training on it outperforms expert execution traces, suggesting that understanding environments is more fundamental than imitation.

\end{abstract}

\section{Introduction}

\begin{wrapfigure}{r}{0.5\textwidth}
    \vspace{-10pt}
    \centering
    \small
    \includegraphics[width=0.5\textwidth]{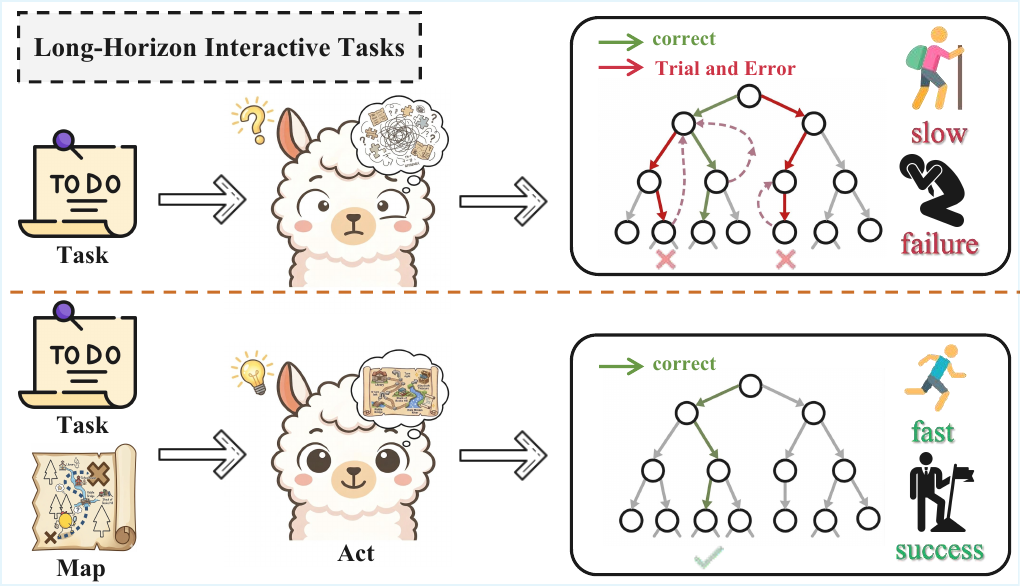}
            \vspace{-18pt}
    \caption{Traditional act-during-think (top) vs. Our map-then-act paradigm (bottom).}
    \label{fig:daemon_vis}
    \vspace{-9pt}
\end{wrapfigure}
Large Language Models (LLMs) have rapidly evolved into autonomous agents capable of long-horizon goal completion~\citep{gur2023real, ho2020constructing, shi2025tool}. Current mainstream paradigms, such as ReAct~\citep{yao2022react} and Chain-of-Thought (CoT)~\citep{zhang2022automatic}, primarily follow a goal-conditioned stepwise planning framework: the agent reasons over the current observation and immediately selects the next action. Existing progress has largely focused on two directions to optimize this cycle: improving reasoning capability through expert trajectories, parameter optimization, or experience replay~\citep{luo2025empirical, ouyang2022training, shi2025continual}; and enhancing memory systems through external trajectory storage or distilled knowledge retrieval to augment decision-making context~\citep{fu2024autoguide, shinn2023reflexion, xia2026skillrl, xu2025mem}. Despite their differences, these approaches share a common structural limitation: environmental understanding is coupled with task execution, acquired reactively as a byproduct of acting.

We term this limitation \textbf{Delayed Environmental Perception}. In 
existing paradigms, agents are forced into a temporal inversion 
where they must ``act to understand''---inferring spatial layouts, 
object-action affordances, and latent constraints only through 
trial-and-error feedback. Crucially, this is a paradigm-level 
bottleneck that cannot be resolved by scaling reasoning capabilities 
alone: a more capable model operating under the same paradigm still 
perceives the environment only as a byproduct of acting within it. 
The recently released ARC-AGI-3 benchmark~\citep{foundation2026arc} 
provides compelling evidence of this limitation---even frontier 
models such as Claude 4.6 achieve near-zero performance in its 
zero-knowledge interactive environments, confirming that strong 
reasoning becomes effectively ungrounded when environmental 
structure is unknown prior to execution.

This delayed perception directly induces an \textbf{Epistemic 
Bottleneck}: without a proactive environmental understanding, agents fall into two characteristic failure modes---Goal Drift, where they 
become trapped in locally plausible but globally suboptimal 
behaviors, and Redundant Trial-and-Error, where they 
repeatedly attempt actions that violate latent environmental logic.

To address this bottleneck, we first draw inspiration from 
Gibson's Affordance Theory~\citep{gibson2014ecological}, 
which suggests that intelligent organisms do not merely ``infer'' 
environmental constraints through failure; instead, they perceive 
action affordances directly from the spatial layout 
prior to execution. This insight motivates a fundamental 
paradigm shift: explicitly decoupling environmental understanding 
from task execution, establishing a global environmental prior 
before acting rather than acquiring it reactively as a 
byproduct of execution. We capture this shift as a spatial extension 
of the ``Let's think step by step'' principle~\citep{kojima2022large} 
to: \textbf{``Let's look around first''}. To operationalize this paradigm shift into a concrete computational 
framework, we draw on Tolman's Cognitive Map 
Theory~\citep{tolman1948cognitive}, which demonstrates that 
organisms navigate unfamiliar environments by first constructing 
structured internal representations through active exploration, 
rather than relying on simple stimulus-response associations. This 
naturally motivates our proposed \textbf{M}ap-then-\textbf{A}ct 
\textbf{P}aradigm (\textbf{MAP}), which introduces an explicit 
\token{map} phase before action execution: \token{map} $\rightarrow$ 
(\token{think} \text{[observation]} \token{act}). MAP consists of 
three stages: \ding{182} Cross-Task Global Exploration, 
for extracting reusable environment-general priors; \ding{183} 
Task-Specific Cognitive Mapping, for constructing 
structured maps of spatial layouts and object-action affordances; 
and \ding{184} Knowledge-Augmented Execution, where 
actions are grounded on the self-generated map rather than raw 
observations alone.

We evaluate MAP across diverse interactive benchmarks, including ALFWorld, TextCraft, ScienceWorld, and the ARC-AGI-3 benchmark for fluid intelligence in fully novel environments. Results show two key findings: \ding{182} MAP consistently improves success rates while reducing interaction steps across tasks without parameter updates; and \ding{183} after lightweight fine-tuning on \textbf{MAP-2K}, a compact dataset of map-then-act trajectories, the resulting \textbf{MAP-4B} substantially outperforms counterparts trained on traditional expert execution traces. These results suggest that teaching agents to understand environments is more fundamental than teaching them to imitate solutions.

\section{Related Work}

\subsection{LLM Agents in Long-Horizon Tasks}

Recent advances in LLMs have spurred growing interest in building agents for complex, long-horizon tasks~\citep{fang2025comprehensive,jimenez2023swe,wang2023voyager}. Early prompting-based agents are prone to planning hallucinations and brainless trial-and-error due to their static workflows. Optimization of these methods has focused on two aspects. For reasoning capability, data-driven methods improve decision-making via imitation learning on expert trajectories, while RL-based methods learn from environment interactions through experience replay~\cite{chen2026learning,peng2025tree,wei2026agentic,xinjie2025reagent}. For memory management, some works store past trajectories or distill interaction history into external memory to augment decision-making context~\citep{park2023generative,shi2025look,yang2025learning,yang2025coarse}, while others maintain skill libraries that retrieve reusable action primitives or task-specific knowledge to guide execution~\citep{asai2023self,cao2025remember,mi2026procmem,zhai2025agentevolver}. However, these approaches remain focused on execution logic and are heavily dependent on external resources, leaving a critical gap in how agents perceive and model the environment itself. In contrast, our proposed MAP emphasizes autonomous environment understanding through self-directed exploration, reducing reliance on external resources and enabling the agent to grow through its own experience.

\subsection{Environment Understanding}

Effective environment-aware task execution requires agents to maintain an accurate understanding of the environment~\citep{liu2026exploratory,yuksekgonul2026learning,zhang2025agent}.
Recent studies~\citep{foundation2026arc,liu2026llm} show that many existing agents operate in a "blind execution" regime, where failures stem not from limited reasoning ability but from insufficient modeling of the environment's underlying structure. Even when succeeding via trial-and-error or imitation, agents often fail to capture fundamental properties such as spatial layouts and object affordances, suggesting a lack of structured environment representation.

Existing memory mechanisms---such as long-context windows or key-value memory modules~\cite{xia2025experience,zhang2025g,zhang2025memgen,zhou2025mem1}---struggle to form consistent environmental models, as fragmented interaction histories are difficult to organize into structured spatial or physical representations. Related efforts in model-based reinforcement learning aim to learn environment dynamics for planning, but typically rely on parametric simulators or latent dynamics models, making them less compatible with language-based agents and open-ended environments. In contrast, evidence from the VLM domain suggests that explicitly modeling spatial structure and viewpoints improves reasoning performance~\citep{wang2025efficient,yin2025spatial}. Motivated by this, MAP introduces a dedicated mapping stage that constructs a cognitive map $M_t$ capturing spatial layouts and object-action affordances prior to execution.

\section{Method}
\label{sec:method}

In this section, we introduce \textbf{MAP}, which enhances the LLM agent's performance by decoupling autonomous environment understanding from task execution. We first formalize the "map-then-act" paradigm as a principled alternative to conventional "act-during-think" baselines (§\ref{sec:task_formulation}). Building on this, we describe our three-stage architecture for transforming environmental interactions into structured cognitive maps (§\ref{sec:architecture}). Finally, we introduce an exploration-driven fine-tuning strategy to internalize these capabilities, demonstrating that distilling map-then-act trajectories is more foundational for generalization than mimicking expert execution (§\ref{sec:finetuning}).

\subsection{Task Formulation}
\label{sec:task_formulation}

Our work enhances agent generalization by explicitly decoupling environmental understanding from task execution. We represent the final execution trajectory as $e = (u, a_1, o_1, \ldots, a_n)$, where $u \in \mathcal{U}$ is the task instruction, $a \in \mathcal{A}$ is the agent action, and $o \in \mathcal{O}$ is the environmental observation.

In the standard ``Act-during-Think'' paradigm, the agent $\pi_\theta$ generates actions conditioned solely on the task instruction and interaction history:
\begin{equation}
    \pi_\theta(e \mid u) = \prod_{t=1}^{n} 
    \pi_\theta(a_t \mid u, a_1, o_1, \ldots, o_{t-1}).
    \label{eq:implicit}
\end{equation}
This formulation is fundamentally constrained to the observational distribution $P(a_t \mid o_t)$---learning \emph{what actions to take} under given observations, but never estimating \emph{how the environment responds}. By Pearl's do-calculus~\citep{pearl2022causal}, observational data alone cannot recover the interventional distribution $P(o_{t+1} \mid do(a_t))$, which is the formal root of the Epistemic Bottleneck.

To address this, MAP divides the workflow of $\pi_\theta$ into two stages. In the mapping stage, the agent actively probes the environment via $do(\text{explore})$, generating exploration trajectories $\tau_{\text{exp}}$ and distilling them into a cognitive map $M$ that encodes causally grounded environmental knowledge:
\begin{equation}
    M \sim \pi_\theta(M \mid u, \tau_{\text{exp}}).
    \label{eq:mapping}
\end{equation}

In the acting stage, $\pi_\theta$ completes the task conditioned on $M$:
\begin{equation}
    \pi_\theta(e \mid u, M) = \prod_{t=1}^{n} 
    \pi_\theta(a_t \mid u, M, a_1, o_1, \ldots, o_{t-1}) 
    \cdot \, \pi_\theta(M \mid u, \tau_{\text{exp}}).
    \label{eq:map_act}
\end{equation}
By conditioning on $M$---a structured summary of interventional experience $do(\text{explore})$---rather than observational history alone, MAP transitions the agent from correlational pattern-matching to causally grounded, knowledge-driven reasoning.

\begin{figure}[h]
\centering
\includegraphics[width=\columnwidth]{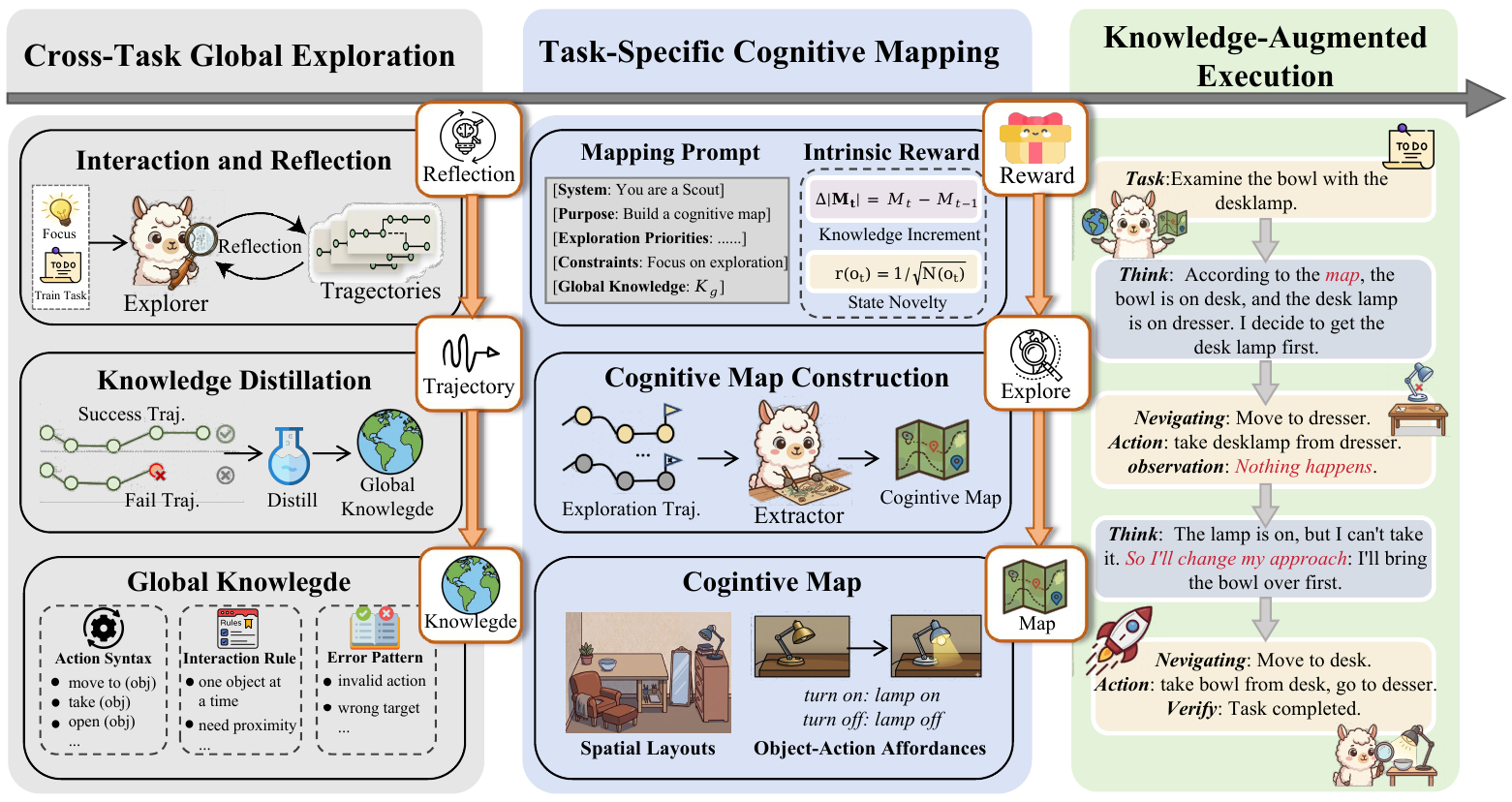}
\caption{Overview of the MAP framework. (\textbf{Left}) \textbf{Cross-Task Global Exploration}: The agent explores training environments to distill cross-task priors $K_g$ capturing general operational rules. (\textbf{Middle}) \textbf{Task-Specific Cognitive Mapping}: Guided by $K_g$ and an RPP prompt protocol, the agent constructs a cognitive map $M_t$ encoding spatial layouts and object-action affordances for the current task. (\textbf{Right}) \textbf{Knowledge-Augmented Execution}: The agent completes the task conditioned on $\{K_g, M_t\}$, bypassing the Epistemic Bottleneck of 
conventional stepwise paradigms.}
\label{fig:method}
\vspace{-1em}
\end{figure}

\subsection{MAP Architecture}
\label{sec:architecture}

In this section, we present the three-stage architecture of MAP. The mapping stage is further decomposed into two lightweight sub-stages, resulting in the three-stage pipeline illustrated in Figure~\ref{fig:method}.

\subsubsection{Cross-Task Global Exploration}
\label{sec:stage1}

The goal of this stage is to discover environment-level general rules 
shared across all tasks, including action syntax, interaction rules, 
and error patterns, independent of specific task goals. This stage 
is executed once per environment and produces a persistent knowledge 
base $K_g$ reused across all subsequent task instances.

\paragraph{Exploration Protocol.}
Taking $Desc_{env}$ and a small set of manual trajectories $F_{manual}$ 
as input, the agent $\pi_\theta$ first acts as a \textit{Focus Analyzer} to derive 
Focus Points ($FP$): actionable exploration priorities that 
guide the investigation of interaction patterns, constraints, and 
conventions (e.g., \textit{``Probe whether the environment enforces 
strict action syntax by testing different command formats and observing 
which are accepted or rejected''}). Guided by $FP$, the agent then 
acts as an \textit{Explorer} on $T_{train}$, executing multiple rounds 
of ``think-act'' iterations. Any failure triggers a \textit{Reflector} 
to perform introspective reflection, with insights incorporated into 
task-specific reflections $\nu$ to assist subsequent retry attempts. 
The resulting trajectories $\tau_{\text{exp}}$, encompassing both 
successful and failed interactions, are passed to the knowledge 
distillation phase.

\paragraph{Knowledge Distillation.}
The agent $\pi_\theta$ distills $\tau_{\text{exp}} = (a_1, o_1, 
\ldots, a_N, o_N)$ into structured environment-general rules $K_g$:
\begin{equation}
    K_g = f_{\text{distill}}\!\left( \tau_{\text{exp}}^{(1)}, 
    \tau_{\text{exp}}^{(2)}, \ldots, \tau_{\text{exp}}^{(N)} \right),
    \label{eq:global_knowledge}
\end{equation}
where $f_{\text{distill}}$ extracts universal patterns from actions 
$a_t$ and observations $o_t$ (details in 
Appendix~\ref{app:prompt_distill}), organized as follows:

\begin{promptbox}{Structure of Cross-Task Global Knowledge}
\it
\begin{itemize}[leftmargin=*, after=\vspace{-0.5em}, itemsep=1pt]
    \item \textbf{Action Syntax:} valid action commands and their 
    exact format.
    \item \textbf{Interaction Rules:} preconditions and constraints for valid action execution.
    \item \textbf{Error Patterns:} recurring failure modes and 
    their diagnostic signals.
\end{itemize}
\end{promptbox}

\noindent Once constructed, $K_g$ serves as a persistent cognitive 
prior injected into the system context for all downstream task 
instances, allowing agents to bypass redundant rule verification 
and focus on task-specific uncertainties from the outset.

\subsubsection{Task-Specific Cognitive Mapping}
\label{sec:stage2}

Guided by the global prior $K_g$, this stage constructs a 
task-specific cognitive map $M_t$ by acquiring concrete facts 
regarding spatial layouts, environmental physics, and object-action 
affordances tailored to the current environment instance.

\paragraph{Adaptive Exploration.}

We define an intrinsic reward $r_{\text{intrinsic}}$ to quantify 
information gain and reduce epistemic uncertainty about the task 
goal $g$, consisting of two metrics:

\begin{itemize}[leftmargin=*]
    \item \textbf{Knowledge Increment (Cond\_A):} $\Delta|M_t| = 
    |M_t| - |M_{t-1}|$, where $|M_t|$ denotes the number of 
    distinct knowledge entries (e.g., confirmed object locations, 
    discovered affordances) at step $t$. A positive increment 
    indicates the discovery of new spatial or relational facts; 
    convergence is declared when $\Delta|M_t| = 0$ persists for 
    $k$ consecutive steps.
    \item \textbf{State Novelty (Cond\_B):} $r(o_t) = 1 / 
    \sqrt{N(o_t)}$, where $N(o_t)$ is the visit count of 
    observation $o_t$. This reward decays as states are revisited, 
    incentivizing the agent to explore unvisited regions. 
    Convergence is declared when $r(o_t)$ falls below threshold 
    $\epsilon$ for $k$ consecutive steps.
\end{itemize}

\paragraph{Dual-Convergence Stopping Criterion.}
The exploration horizon is dynamically determined by:
\begin{equation}
    T_{\text{stop}} = \min \bigl\{ t \mid (\text{Cond\_A}_t \wedge 
    \text{Cond\_B}_t \text{ converge}) \;\vee\; (t \geq T_{\max}) 
    \bigr\}.
    \label{eq:t_stop}
\end{equation}
Both conditions must converge simultaneously: Cond\_A ensures map 
completeness, while Cond\_B ensures exploration diversity. Requiring 
both prevents premature termination, as an agent may stop discovering 
new facts while still traversing novel regions, or vice versa. 
A detailed analysis is provided in Appendix~\ref{app:stopping_criterion}.

\paragraph{Task Mapping Prompt Skeleton.}
We design a structured Role-Purpose-Priority (RPP) protocol 
to guide systematic environmental mapping. Prompt skeletons are provided in Appendix~\ref{app:prompt_mapping}.

\paragraph{Cognitive Map Construction.}
Upon triggering the stop signal, a Key Information Extractor performs 
structured analysis of $\tau_{\text{exp}}$ to generate $M_t$:
\begin{equation}
    M_t = f_{\text{map}}\!\left(\tau_{\text{exp}},\;\;u \right).
    \label{eq:task_map}
\end{equation}

\begin{promptbox}{Structure of Task-Specific Cognitive Map}
\it
\begin{itemize}[leftmargin=*, after=\vspace{-0.5em}, itemsep=1pt]
    \item \textbf{Spatial Layouts:} the structural organization of 
    the environment, including reachable regions, object locations, 
    and their spatial relationships.
    \item \textbf{Object-Action Affordances:} interactable entities 
    and the effects their associated actions produce on the 
    environment.
    \item \textbf{Game Rules} \textit{(ARC-AGI-3):} 
    environment-specific mechanics and latent rules governing 
    valid interactions.
\end{itemize}
\end{promptbox}

\subsubsection{Knowledge-Augmented Execution}
\label{sec:stage3}

In the final execution stage, the agent applies the dual-layer framework—comprising the global prior $K_g$ and the task-specific cognitive map $M_t$—enabling proactive, knowledge-driven reasoning. Specifically, at time step $t$, the action $a_t$ is sampled conditioned on the task instruction $u$, the cognitive map $M_t$, the global prior $K_g$, and the interaction history $h_t = (a_1, o_1, \dots, a_{t-1}, o_{t-1})$:
\begin{equation}
    a_t \sim \pi_\theta(a_t \mid u, M_t, K_g, h_t).
    \label{eq:execution_policy}
\end{equation}

\subsection{Internalization via Cognitive Fine-tuning}
\label{sec:finetuning}

While inference-time prompting demonstrates the effectiveness of MAP, 
we further investigate whether such environment-understanding 
capabilities can be internalized into model parameters. To this end, 
we propose a teacher-student distillation pipeline to construct 
\textbf{MAP-2K}, where state-of-the-art LLMs (e.g., GPT-4.1, Claude 
4.5) execute the MAP pipeline as expert annotators, 
generating full map-then-act trajectories $\tau_{\text{MAP}}$ given task instruction $u$:
\begin{equation}
    \tau_{\text{MAP}} = f_{\text{teacher}}(u).
\end{equation}
To ensure fidelity, the synthetic trajectories undergo a rigorous 
ground-truth alignment check against the environment engine's 
internal state to correct potential hallucinations.

We then fine-tune the student model $\pi_\theta$ on MAP-2K. For a 
map-then-act trajectory $\tau_{\text{MAP}} = (a_1, o_1, \ldots, a_N, o_N)$, 
we minimize:
\begin{equation}
    \mathcal{L}_{\text{MAP}} = -\sum_{t=1}^{N} \log \pi_\theta(a_t 
    \mid o_{<t}, a_{<t}),
\end{equation}
where $\pi_\theta$ is the LLM policy being trained. The loss 
supervises the full action sequence across stages, 
directly internalizing both the environment-understanding and 
task-execution capabilities into $\pi_\theta$. Unlike traditional 
tuning that supervises on expert execution traces alone, MAP-2K 
trains the agent on complete map-then-act trajectories, teaching 
it to first understand the environment through active exploration 
and then ground its decisions in structured knowledge, rather than 
merely memorizing \textit{what} actions to take.

\section{Experiment}

In this paper, we conduct experiments to answer the following research questions (RQs):

\textbf{RQ1}: Does MAP consistently outperform existing agent paradigms across benchmarks, and does MAP-2K offer superior training signal over expert execution trajectories?

\textbf{RQ2}: Does Mapping enable agents to develop genuine causal understanding of the environment?

\textbf{RQ3}: Is the exploration overhead of MAP's mapping phase computationally acceptable?

\textbf{RQ4}: Is each stage of MAP individually necessary?

\subsection{Experimental Setups}
\label{sec:exp_setup}
\paragraph{Environments.}
We evaluate on four benchmarks: \ding{192} ALFWorld~\citep{
shridhar2020alfworld}, a household task environment requiring 
navigation and object manipulation; \ding{193} TextCraft~
\citep{sanghi2022textcraft}, a Minecraft-inspired crafting 
environment with multi-step recipes; \ding{194} ScienceWorld~
\citep{wang2022scienceworld}, a text-based science task benchmark 
requiring procedural reasoning; and \ding{195} ARC-AGI-3~
\citep{foundation2026arc}, a game benchmark of abstract 
turn-based environments with no explicit rules or goals. Detailed 
descriptions are provided in Appendix~\ref{app:environments}.

\paragraph{Implementation details.}
To ensure robust evaluation, our training data MAP-2K underwent strict decontamination to prevent repository-level overlap with benchmarks. We fine-tuned the Qwen3-4B-Thinking model~\citep{yang2025qwen3} using ms-swift on 8 NVIDIA H800 GPUs which referred to as MAP-4B. The learning rate was set to $1 \times 10^{-5}$, and training was conducted for 3 epochs.
To ensure fair comparison, we constrain all baselines to the same total step budget as MAP.

\paragraph{LLM Backbones.}
We evaluated a diverse array of models, including Claude, GPT~\citep{achiam2023gpt}, Kimi~\citep{team2025kimi}, Minimax~\citep{chen2025minimax}, Doubao, Deepseek and Qwen~\citep{yang2025qwen3} series. 

\paragraph{Baselines.}
We compare MAP against three established paradigms: \ding{192} Standard ReAct: A goal-driven stepwise planning framework interleaving reasoning and action. \ding{193} Map-and-Act (CoMAP): A non-staged variant where environmental mapping and task execution are performed simultaneously (detailed in Appendix~\ref{app:comap_prompt}). \ding{194} SFT-Execution (ACT-4B): To ensure a fair comparison, we fine-tuned Qwen3-4B-Thinking on 2K expert execution trajectories using the same configurations as MAP-4B.

\begin{table*}[t]
\centering
\caption{Performance comparison on long-horizon interactive benchmarks. \upred{n} and 
\downblue{n} indicate performance improvements and degradations relative to 
the preceding paradigm (CoMAP vs. ReAct and MAP vs. CoMAP).}
\resizebox{1\textwidth}{!}{
\begin{tabular}{lccccccccc}
\toprule
\multirow{2}{*}{\textbf{Model}} 
& \multicolumn{3}{c}{\textbf{Alfworld}} 
& \multicolumn{3}{c}{\textbf{Textcraft}} 
& \multicolumn{3}{c}{\textbf{Sciworld}} \\
\cmidrule(lr){2-4} \cmidrule(lr){5-7} \cmidrule(lr){8-10}
& \textbf{ReAct} & \textbf{CoMAP} & \textbf{MAP}
& \textbf{ReAct} & \textbf{CoMAP} & \textbf{MAP}
& \textbf{ReAct} & \textbf{CoMAP} & \textbf{MAP} \\
\midrule
\multicolumn{10}{c}{\textbf{Non-reasoning Models}}\\
\midrule
\lightgraybg Qwen3-4B-Instruct 
& 31.8 & 39.0\upred{7.2} & 61.1\upred{22.1} 
& 31.8 & 37.1\upred{5.3} & 71.9\upred{34.8} 
& 13.9 & 16.3\upred{2.4} & 24.3\upred{8.0} \\

Qwen3-30B-A3B-Instruct 
& 39.3 & 52.3\upred{13.0} & 62.0\upred{9.7} 
& 56.8 & 58.6\upred{1.8} & 66.6\upred{8.0} 
& 0.5 & 1.5\upred{1.0} & 9.8\upred{8.3} \\

\lightgraybg GPT-4o-0806 
& 76.1 & 82.3\upred{6.2} & 86.6\upred{4.3} 
& 91.6 & 92.5\upred{0.9} & 96.7\upred{4.2} 
& 31.4 & 32.7\upred{1.3} & 38.2\upred{5.5} \\

GPT-4o-1120 
& 75.2 & 79.1\upred{3.9} & 82.6\upred{3.5} 
& 94.7 & 96.7\upred{2.0} & 99.6\upred{2.9} 
& 41.3 & 44.8\upred{3.5} & 53.3\upred{8.5} \\

\lightgraybg GPT-4o-mini 
& 21.9 & 31.0\upred{9.1} & 46.2\upred{15.2} 
& 46.5 & 54.2\upred{7.7} & 75.6\upred{21.4} 
& 1.6 & 6.2\upred{4.6} & 12.2\upred{6.0} \\

\midrule
\multicolumn{10}{c}{\textbf{Reasoning Models}}\\
\midrule
\lightgraybg Qwen3-4B-Thinking 
& 58.5 & 61.9\upred{3.4} & 71.5\upred{9.6} 
& 35.6 & 37.5\upred{1.9} & 59.6\upred{22.1} 
& 1.5 & 9.2\upred{7.7} & 11.4\upred{2.2} \\

Qwen3-8B-Thinking 
& 69.3 & 74.6\upred{5.3} & 78.0\upred{3.4} 
& 66.4 & 67.0\upred{0.6} & 87.1\upred{20.1} 
& 25.8 & 28.7\upred{2.9} & 32.4\upred{3.7} \\

\lightgraybg Qwen3-32B-Thinking 
& 78.4 & 80.7\upred{2.3} & 84.0\upred{3.3} 
& 82.4 & 87.7\upred{5.3} & 90.4\upred{2.7} 
& 16.1 & 19.7\upred{3.6} & 28.0\upred{8.3} \\

Kimi-K2-Thinking 
& 89.1 & 91.1\upred{2.0} & 95.0\upred{3.9} 
& 96.8 & 97.6\upred{0.8} & 99.2\upred{1.6} 
& 41.2 & 43.5\upred{2.3} & 46.8\upred{3.3} \\

\lightgraybg Minimax-M2.7 
& 84.5 & 87.9\upred{3.4} & 88.1\upred{0.2}
& 92.3 & 92.1\downblue{0.2} & 96.7\upred{4.6} 
& 51.1 & 50.2\downblue{0.9} & 54.2\upred{4.0} \\

Deepseek-V4-Pro
& 96.6 & 97.2\upred{0.6} & 99.6\upred{2.4}
& 98.8 & 99.2\upred{0.4} & 99.5\upred{0.3}
& 32.3 & 36.8\upred{4.5} & 41.2\upred{4.4} \\

\lightgraybg Doubao-Seed-2.0-Pro 
& 93.1 & 94.5\upred{1.4} & 96.3\upred{1.8}
& 63.1 & 65.6\upred{2.5} & 69.8\upred{4.2}
& 42.5 & 44.1\upred{1.6} & 48.9\upred{4.8} \\

\midrule
\multicolumn{10}{c}{\textbf{Fine-Tuning Models}}\\
\midrule
ACT-4B 
& 78.2 & 80.7\upred{2.5} & 84.3\upred{3.6}
& 70.1 & 74.5\upred{4.4} & 79.4\upred{4.9}
& 16.9 & 20.1\upred{3.2} & 23.6\upred{3.5} \\

\graybg \textbf{MAP-4B} 
& 87.1 & 85.6\downblue{1.5} & 94.1\upred{8.5}
& 91.2 & 92.3\upred{1.1} & 95.6\upred{3.3}
& 35.0 & 37.2\upred{2.2} & 40.5\upred{3.3} \\

\bottomrule
\end{tabular}
}
\label{tab:main_results}
\vspace{-0.5em}
\end{table*}

\newcolumntype{q}{>{\columncolor{blue!6}}c}

\begin{table}[t]
\centering
\small
\renewcommand{\arraystretch}{0.9}
\caption{Performance comparison on ARC-AGI-3 benchmarks. We report 
both achieved level and success score on Claude 4.6 Opus.}
\vspace{-0.2cm}
\resizebox{0.98\textwidth}{!}{
\begin{tabular}{l *{6}{cc}}
\toprule
\multirow{2}{*}{\textbf{Method}}
& \multicolumn{2}{c}{\textbf{TU93}}
& \multicolumn{2}{c}{\textbf{SB26}}
& \multicolumn{2}{c}{\textbf{VC33}}
& \multicolumn{2}{c}{\textbf{RE86}}
& \multicolumn{2}{c}{\textbf{AR25}}
& \multicolumn{2}{c}{\textbf{WA30}} \\

\cmidrule(lr){2-3} \cmidrule(lr){4-5} \cmidrule(lr){6-7}
\cmidrule(lr){8-9} \cmidrule(lr){10-11} \cmidrule(lr){12-13}

& Level & Score
& Level & Score
& Level & Score
& Level & Score
& Level & Score
& Level & Score \\

\midrule

ReAct
& 0 & 0.00
& 1 & 0.19
& 0 & 0.00
& 0 & 0.00
& 0 & 0.00
& 0 & 0.00 \\

MAP
& \cellcolor{blue!6}\textbf{4} & \cellcolor{blue!6}\textbf{3.34}
& \cellcolor{blue!6}\textbf{3} & \cellcolor{blue!6}\textbf{7.59}
& \cellcolor{blue!6}\textbf{3} & \cellcolor{blue!6}\textbf{4.12}
& \cellcolor{blue!6}\textbf{3} & \cellcolor{blue!6}\textbf{11.59}
& \cellcolor{blue!6}\textbf{3} & \cellcolor{blue!6}\textbf{7.66}
& \cellcolor{blue!6}\textbf{2} & \cellcolor{blue!6}\textbf{6.67} \\

\bottomrule
\end{tabular}
}
\label{tab:arc_results}
\vspace{-0.8em}
\end{table}

\subsection{Main Results (RQ1)}

We evaluate MAP on two types of benchmarks: \textbf{(1) long-horizon 
interactive benchmarks} (ALFWorld, TextCraft, and ScienceWorld) that 
test task completion in structured but unfamiliar environments; and 
\textbf{(2) fluid intelligence benchmarks} (ARC-AGI-3) that test 
adaptation and rule discovery in fully novel environments.

\subsubsection{Results on Long-Horizon Interactive Benchmarks}

Table~\ref{tab:main_results} summarizes the performance of MAP and baselines across 
ALFWorld, TextCraft, and ScienceWorld. Our analysis reveals two key findings.

\noindent\textbf{Environmental understanding is critical, and staged 
decoupling further amplifies its benefit.}
Across most benchmarks and backbones, performance follows a consistent 
ordering under comparable token budgets: ReAct $<$ CoMAP $<$ MAP. 
CoMAP already improves over ReAct, confirming that environmental 
understanding is essential. However, it consistently falls below MAP, 
indicating that both \emph{when} and \emph{how} cognitive mapping is 
performed matter. By separating mapping from execution, MAP enables 
agents to build a coherent cognitive map before acting.

\noindent\textbf{MAP-2K provides superior training signal over expert execution trajectories.}
Under identical training settings, MAP-4B substantially outperforms 
ACT-4B across all benchmarks and surpasses several larger models, 
validating MAP-2K as a high-quality training source. Unlike expert 
execution traces, map-then-act trajectories capture environmental 
understanding rather than surface-level action imitation, suggesting 
that \textit{teaching agents to understand environments is more 
foundational than teaching them to complete tasks}.

\subsubsection{Results on Fluid Intelligence Benchmarks}

We further evaluate MAP on ARC-AGI-3, where agents must explore 
unknown game worlds without any explicit rules or goals. We adopt 
Claude 4.6 Opus as the backbone, as it represents the current 
state-of-the-art in reasoning and execution capability. Under the 
standard \textit{ReAct} framework, performance remains near-zero 
across all environments, highlighting the fundamental challenge 
this benchmark poses to conventional paradigms. In contrast, MAP 
achieves consistent improvements across 22 out of 25 games 
(Table~\ref{tab:arc_results}; full results in 
Appendix~\ref{app:arc_full_results}), demonstrating broad 
generalization to previously unseen environments. These results 
confirm that the bottleneck lies not in reasoning capability, but 
in the absence of explicit environmental understanding---a gap 
that MAP directly addresses through structured pre-execution 
exploration.

\subsection{Environmental Understanding Ability (RQ2)}
\begin{wrapfigure}{r}{0.33\textwidth}
    \centering
    \vspace{-2em}
    \includegraphics[width=\linewidth]{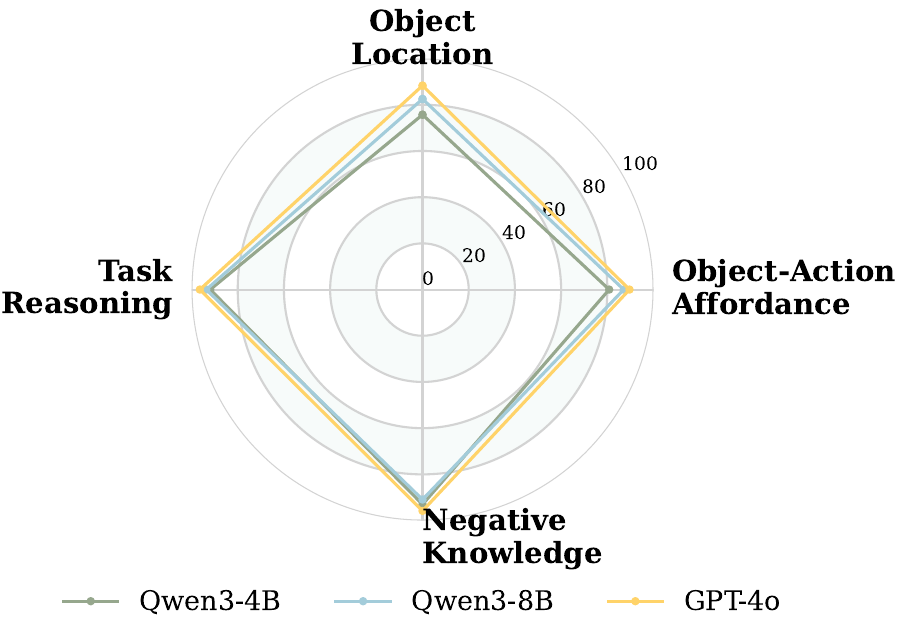}
    \vspace{-10pt}
    \caption{Map QA accuracy evaluated on ALFWorld's constructed cognitive maps $M_t$.}
    \label{fig:qa_radar}
    \vspace{-10pt}
\end{wrapfigure}

To answer whether the mapping stage enables agents to develop genuine 
causal understanding of the environment, we design three complementary experiments.

\noindent\textbf{Map QA Accuracy.}
We design an offline QA evaluation covering four categories of 
environment-probing questions (detailed in Appendix~\ref{app:map_qa}): \emph{object locations}, \emph{object-action affordance}, \emph{negative knowledge}, and \emph{task reasoning}. The agent is queried 
solely based on its constructed $M_t$ and evaluated against ground-truth 
states extracted from the environment engine. As shown in 
Figure~\ref{fig:qa_radar}, all models achieve strong accuracy across all 
four categories, confirming that $M_t$ faithfully captures the underlying structure of the environment prior to execution.

\noindent\textbf{Rule Discovery in Novel Environments.}
We further probe whether structured exploration enables agents to 
discover underlying rules in completely unknown environments, using 
ARC-AGI-3 as a testbed. Unlike the previous two experiments where 
environment rules are predefined, ARC-AGI-3 provides no explicit rules 
or goals---agents must autonomously infer the world's underlying logic 
through interaction. MAP enables Claude 4.6 Opus to progressively advance 
through multiple levels: by systematically mapping the game environment, 
the agent constructs a structured understanding of the underlying game 
mechanics, which in turn guides informed decision-making to drive game 
progression. Details are provided in Appendix~\ref{app:case_arcagi3}.

\begin{wraptable}{r}{0.55\textwidth}
    \centering
    \vspace{-12pt}
    \caption{Comparison under dynamic perturbation on ALFWorld. Base denotes Qwen3-4B-Thinking without fine-tuning.}
    \resizebox{0.55\textwidth}{!}{
        \begin{tabular}{>{\raggedright\arraybackslash}lcccc}
        \toprule
        \textbf{Model} 
        & \textbf{pass@1} 
        & \textbf{pass@1$_{\text{perturb}}$}
        & \textbf{Re-explore Rate(\%) $\uparrow$} 
        & \textbf{$\Delta$Steps $\downarrow$} \\
        \midrule
        Base & 58.5 & 45.6\downblue{13.3} & 62.8 & 11.2 \\
        ACT-4B & 78.2 & 64.8\downblue{13.4} & 69.0 & 13.0 \\
        \graybg MAP-4B & 87.1 & 79.5\downblue{7.6} & 80.9 & 7.3 \\
        \bottomrule
        \end{tabular}
    }
    \label{tab:perturbation}
    \vspace{-8pt}
\end{wraptable}
\noindent\textbf{Causal Adaptability under Environment Shift.}
We introduce controlled mid-episode perturbations by relocating target objects at a random step, to test whether mapping capability enables adaptive replanning rather than relying on memorized action sequences. As shown in Table~\ref{tab:perturbation}, the untuned base model 
suffers the largest performance drop under perturbation, highlighting 
the brittleness of conventional paradigms. Among fine-tuned models, 
MAP-4B exhibits a substantially smaller drop and recovers more efficiently, consistent with directed re-exploration rather than exhaustive trial-and-error. Detailed 
metric definitions are provided in Appendix~\ref{app:perturbation}.

\subsection{Computational Efficiency Analysis (RQ3)}
\label{sec:computation}

\begin{figure*}
    \centering
    \includegraphics[width=0.9\linewidth]{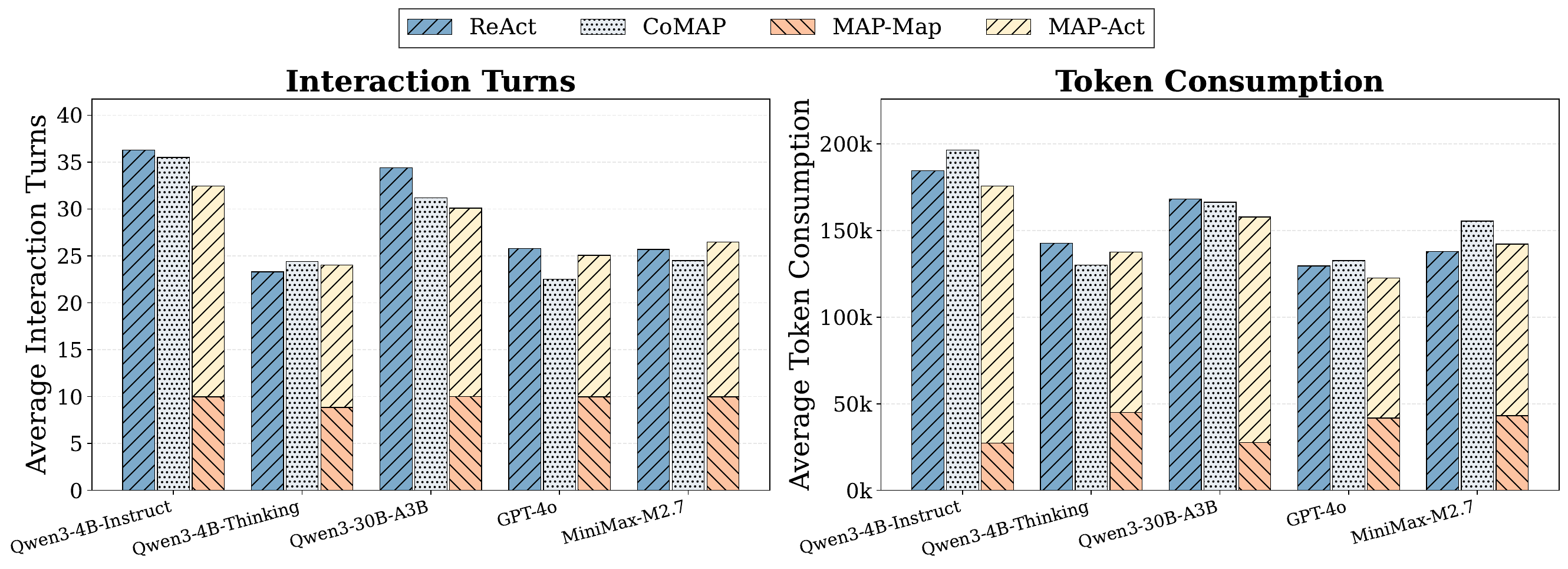}
    \caption{Comparison of interaction turns (left) and token consumption (right) on ALFWorld. Despite upfront mapping overhead, MAP achieves comparable or lower interaction and token cost during execution, indicating that pre-exploration reduces redundancy without additional overhead.}
    \label{fig:efficiency}
\end{figure*}

Under an equivalent total step budget (mapping steps + acting steps), MAP 
consistently achieves higher task success rates---results are reported in 
Table~\ref{tab:main_results}. We further analyze two key metrics: interaction 
turns and token consumption. As shown in Figure~\ref{fig:efficiency}, while 
the mapping phase introduces upfront overhead, it pays dividends in execution 
by eliminating redundant trial-and-error, yielding total costs comparable to 
or lower than ReAct. These results confirm that MAP achieves superior 
Epistemic Efficiency: redistributing the interaction budget toward 
environmental understanding rather than uninformed trial-and-error.

\subsection{Ablation Study (RQ4)}
\label{sec:ablation_study}

To answer whether each stage of MAP is individually necessary, and to 
further validate key design choices, we conduct ablation studies along 
three dimensions on ALFWorld using Qwen3-Thinking models at three scales 
(4B, 8B, and 32B).

\noindent\textbf{Stage Necessity.}
We ablate each of the three stages by removing them individually: 
\textbf{w/o Stage 1} skips global exploration and proceeds directly to 
task mapping; \textbf{w/o Stage 2} removes task mapping entirely, 
injecting only global knowledge $K_g$ into the acting stage. As shown in 
Table~\ref{tab:ablation_stage}, both ablations lead to consistent 
performance drops across all model scales, with w/o Stage 2 incurring the larger degradation, 
confirming that task mapping is the more critical stage while global 
exploration provides complementary gains.

\noindent\textbf{Map Component Necessity.}
To identify which components of the cognitive map $M_t$ are most critical, 
we remove spatial layouts (\textbf{w/o Spatial}) and object-action 
affordances (\textbf{w/o Affordance}) individually. As shown in 
Table~\ref{tab:ablation_map}, removing spatial layouts causes a more 
pronounced drop, suggesting it provides 
the foundational scaffolding for task navigation, while removing 
affordances yields a non-trivial degradation, 
indicating that action consequence modeling contributes independently---
enabling the agent to reason about \emph{what to do} once it knows 
\emph{where to go}. Both components are thus necessary for effective 
decision-making.

\begin{wrapfigure}{r}{0.4\textwidth}
    \centering
    \small
    \vspace{-1.35em}
    \includegraphics[width=0.3\textwidth]{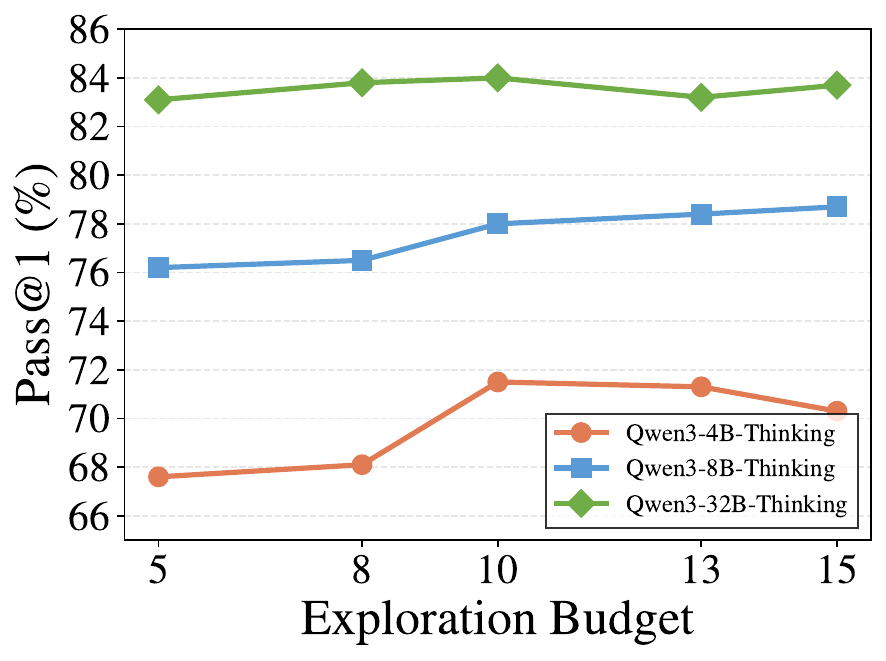}
    \vspace{-10pt}
    \caption{Effect of exploration budget on pass@1 across three model 
    scales on ALFWorld.}
    \label{fig:ablation_budget}
    \vspace{-5pt}
\end{wrapfigure}
\noindent\textbf{Exploration Budget Sensitivity.}
We vary the exploration step budget to 
examine how map quality scales with exploration effort. As shown in 
Figure~\ref{fig:ablation_budget}, performance generally improves with 
more exploration steps and stabilizes beyond 10 steps across all model 
scales. Notably, larger models (32B) are less sensitive to budget 
variations, maintaining strong performance even with minimal exploration, 
while smaller models (4B, 8B) benefit more substantially from additional 
steps. These results suggest that a moderate exploration budget is 
sufficient for MAP to construct an effective cognitive map, without 
requiring exhaustive environment traversal.

\begin{table}[t]
    \centering
    \begin{minipage}{0.48\textwidth}
        \centering
        \small
        \caption{Stage necessity ablation on ALFWorld.}
        \label{tab:ablation_stage}
        \resizebox{\linewidth}{!}{
        \begin{tabular}{>{\raggedright\arraybackslash}lccc}
            \toprule
            \textbf{Setting} & \textbf{4B} & \textbf{8B} & \textbf{32B} \\
            \midrule
            \rowcolor{gray!10} Full Stage & 71.5 & 78.0 & 84.0 \\
            \quad w/o Stage 1 
                & 64.8\downblue{6.7}
                & 73.8\downblue{4.2}
                & 82.3\downblue{1.7} \\
            \quad w/o Stage 2 
                & 55.0\downblue{16.5}
                & 72.1\downblue{5.9}
                & 80.3\downblue{3.7} \\
            \bottomrule
        \end{tabular}
        }
    \end{minipage}
    \hfill
    \begin{minipage}{0.50\textwidth}
        \centering
        \small
        \caption{Map Component ablation on ALFWorld.}
        \label{tab:ablation_map}
        \resizebox{\linewidth}{!}{
        \begin{tabular}{>{\raggedright\arraybackslash}lccc}
            \toprule
            \textbf{Component} & \textbf{4B} & \textbf{8B} & \textbf{32B} \\
            \midrule
            \rowcolor{gray!10} Full Cognitive Map & 71.5 & 78.0 & 84.0 \\
            \quad w/o Spatial 
                & 56.6\downblue{14.9}
                & 74.3\downblue{3.7}
                & 80.0\downblue{4.0} \\
            \quad w/o Affordance 
                & 62.3\downblue{9.2}
                & 76.0\downblue{2.0}
                & 83.3\downblue{0.7} \\
            \bottomrule
        \end{tabular}
        }
    \end{minipage}
    \vspace{-5pt}
\end{table}

\section{Conclusion}

In this work, we identified the fundamental limitation of existing agent 
paradigms---the \textit{epistemic bottleneck}---and proposed MAP, a 
Map-then-Act paradigm that explicitly decouples environmental understanding 
from task execution. Through a structured three-stage pipeline and the 
MAP-2K fine-tuning dataset, MAP consistently outperforms existing 
paradigms across benchmarks and model scales, while MAP-4B surpasses 
models of significantly larger size. Beyond three structured environments, 
MAP enables meaningful progress on ARC-AGI-3 where frontier models 
score near zero, demonstrating that structured exploration is essential 
for agent performance in fully unknown environments. These findings 
suggest that explicit cognitive mapping provides a more robust 
foundation for adaptive, long-horizon interactive agents.

\bibliographystyle{main}   
\newpage
\bibliography{reference}
\newpage
\appendix

\section{Limitations and Future Works} 
\label{app:limit}
While MAP demonstrates strong performance across diverse interactive reasoning benchmarks, the current framework is primarily validated in text-based environments with action spaces. The cognitive mapping mechanism has not yet been extended to embodied AI settings or multimodal perception scenarios, where agents must construct environmental representations from visual inputs and continuous action spaces. We leave the exploration of MAP in embodied and multimodal domains—such as robotic manipulation and vision-language navigation—as an important direction for future work.

\section{More Implementation Details}
\subsection{Dual-Convergence Stopping Criterion Analysis}
\label{app:stopping_criterion}

\begin{wrapfigure}{r}{0.5\textwidth}
    \centering
    \vspace{-10pt}
    \includegraphics[width=0.45\textwidth]{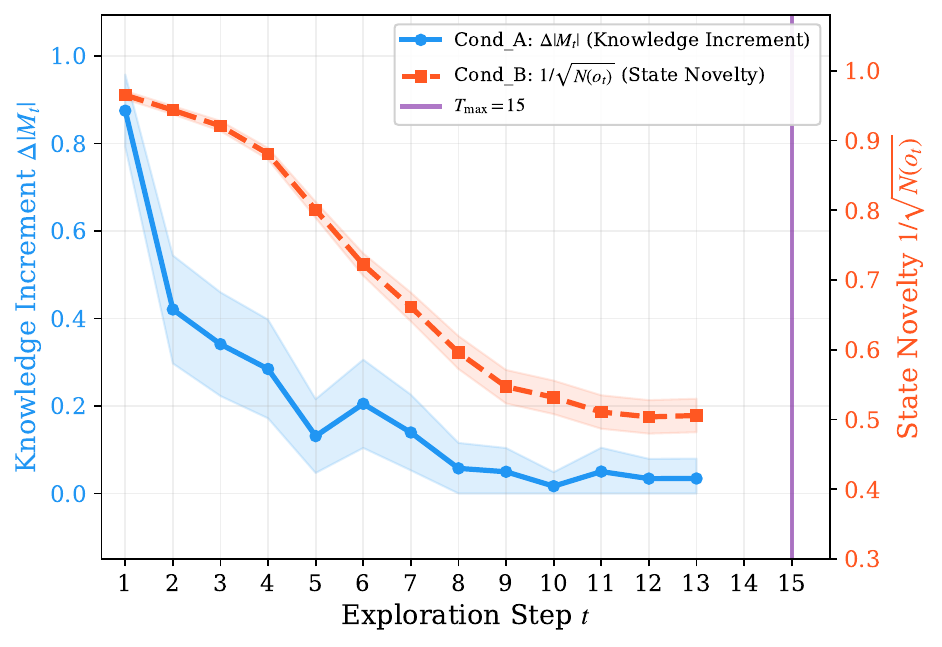}
    \caption{Step-wise dynamics of Knowledge Increment ($\Delta|M_t|$) 
    and State Novelty ($1/\sqrt{N(o_t)}$) during a representative 
    mapping episode on TextCraft.}
    \label{fig:dual_convergence}
    \vspace{-10pt}
\end{wrapfigure}

To validate the dual-convergence stopping criterion empirically, 
we visualize the step-wise dynamics of Knowledge Increment 
($\Delta|M_t|$) and State Novelty ($1/\sqrt{N(o_t)}$) over a 
representative mapping episode on TextCraft 
(Figure~\ref{fig:dual_convergence}). Specifically, Cond\_A is 
satisfied when $\Delta|M_t|$ approaches zero for $W_k=3$ 
consecutive steps, indicating no new spatial or affordance 
information is being discovered; Cond\_B is satisfied when the 
sliding-window average of $r(o_t)$ (window size $W_n=5$) drops 
below $\varepsilon=0.5$, indicating the agent is predominantly 
revisiting previously explored states. A minimum exploration 
floor of $T_{\min}=3$ steps prevents premature termination, 
while $T_{\max}=15$ serves as a safety cap. Both metrics remain 
high in early steps as the agent rapidly acquires new spatial 
layouts and object-action affordances, then jointly converge at 
approximately step 13, at which point $T_{\text{stop}}$ is 
triggered naturally. This confirms that the adaptive stopping 
criterion correctly identifies the natural saturation point of 
the mapping process, avoiding both premature termination and 
redundant over-exploration, thereby allocating the interaction 
budget efficiently across the mapping phase.

\subsection{MAP-2K Dataset Details}
\label{app:dataset}

\begin{wraptable}{r}{0.55\textwidth}
    \centering
    \vspace{-10pt}
    \caption{MAP-2K dataset statistics.}
    \label{tab:dataset_stats}
    \resizebox{0.5\textwidth}{!}{
    \begin{tabular}{lccc}
        \toprule
        \textbf{Environment} & \textbf{\# Traj.} 
        & \textbf{Avg. Steps} & \textbf{Avg. Tokens(k)} \\
        \midrule
        ALFWorld     & 667 & 48.8 & 6.6 \\
        TextCraft    & 667 & 42.9 & 5.7 \\
        ScienceWorld & 667 & 60.7 & 8.2 \\
        \midrule
        \textbf{Total} & $\sim$2,000 & 50.8 & 6.8 \\
        \bottomrule
    \end{tabular}
    }
    \vspace{-5pt}
\end{wraptable}
MAP-2K is constructed via a teacher-student distillation pipeline. We 
employ GPT-4.1 and Claude 4.5 as expert cognitive annotator to execute the MAP 
exploration pipeline across training task instances. For each instance, 
the teacher model is deployed under a goal-free exploration prompt and 
produces an exploration trajectory $\tau_{\text{exp}} = (a_1, o_1, 
\ldots, a_N, o_N)$. To ensure data quality, all synthetic trajectories 
undergo a rigorous ground-truth alignment check against the environment 
engine's internal state to correct potential hallucinations; we filter 
trajectories by requiring them to exceed minimum thresholds on both 
spatial coverage and factual accuracy, yielding a clean distillation 
signal with reduced hallucinated or non-actionable content. The final 
MAP-2K dataset comprises approximately 2,000 expert exploration 
trajectories distributed across three environments, as summarized in 
Table~\ref{tab:dataset_stats}. To prevent test set leakage, exploration 
trajectories are collected exclusively from the training splits of each 
benchmark, following the standard data splits established in the original 
environment papers.

\section{Additional Experimental Details}
\label{app:exp_details}
\subsection{Environment Setup}
\label{app:environments}

We evaluate MAP on three long-horizon interactive benchmarks spanning 
diverse task structures and environmental complexities.

\noindent\textbf{ALFWorld.} ALFWorld is a text-based 
interactive environment that aligns household tasks with a visually rich 
3D simulator, enabling the study of agents capable of both high-level 
planning and grounded execution. The benchmark comprises six task types: 
\textit{Pick \& Place} (locate and move an object to a target location), 
\textit{Examine in Light} (find an object and examine it under a light 
source), \textit{Clean \& Place} (clean an object in a sink and place 
it), \textit{Heat \& Place} (heat an object in a microwave and place 
it), \textit{Cool \& Place} (cool an object in a fridge and place it), 
and \textit{Pick Two \& Place} (locate two objects of the same type and 
move them to a target location). Each task requires a long sequence of 
correct actions across multiple rooms, posing significant challenges for 
sequential decision-making. We evaluate on the 134 unseen test games 
following standard protocol.

\noindent\textbf{TextCraft.} TextCraft is a 
text-based Minecraft-inspired environment where agents must synthesize 
target items through multi-step crafting recipes. Tasks vary in 
complexity from single-step crafting to multi-hop synthesis chains 
requiring the construction of intermediate items, demanding both 
environmental exploration and compositional planning. We follow the 
standard evaluation split of 100 test tasks.

\noindent\textbf{ScienceWorld.} ScienceWorld is a 
text-based virtual environment designed to evaluate scientific reasoning 
and procedural task completion at the level of an elementary school 
science curriculum. The environment features multiple interconnected 
locations (e.g., kitchen, workshop, laboratory, greenhouse) populated 
with over 200 objects possessing diverse physical properties (e.g., 
temperature, conductivity, state of matter). The underlying simulation 
models thermodynamics, electrical circuits, chemical reactions, and 
biological processes, supporting complex state changes and interactions. 
The benchmark comprises 30 task types spanning 10 science topics, 
including changing states of matter, understanding life cycles, and 
building electrical circuits. For each task type, multiple variations 
are procedurally generated to test generalization. We evaluate on the 
standard test split of 200 task instances.

\noindent\textbf{ARC-AGI-3.} ARC-AGI-3 is an 
interactive benchmark designed to evaluate agentic intelligence 
through abstract turn-based game environments structured around a 
64$\times$64 grid with 16 possible colors. Unlike conventional 
benchmarks, agents receive \emph{no explicit rules, goals, or 
instructions}---they must autonomously infer win conditions and 
underlying game mechanics through interaction alone. The benchmark 
evaluates four core capabilities: exploration, world modeling, 
goal inference, and planning under uncertainty. Performance is 
measured via Relative Human Action Efficiency (RHAE), which assesses 
action efficiency relative to human baselines. All environments are 
human-calibrated to ensure 100\% human solvability, yet frontier 
models score below 1\% as of March 2026, making it an ideal testbed 
for evaluating structured exploration under complete environmental 
uncertainty. We evaluate on 6 distinct game environments following 
the standard evaluation protocol.

\noindent\textbf{Step Budget Allocation.} For all environments, the 
total interaction budget is split between the mapping phase (Stage 2) 
and the acting phase (Stage 3). Specifically, we allocate 10 steps for 
mapping and 50 steps for acting in ALFWorld, 15 and 50 steps in 
TextCraft, and 15 and 50 steps in ScienceWorld, respectively.  For 
ARC-AGI-3, given the open-ended nature of the environments and the 
absence of predefined task goals, we allocate 30 steps for the 
mapping phase. The 
global exploration phase (Stage 1) is conducted offline prior to 
evaluation and does not consume the per-episode budget.

\subsection{Full Results on ARC-AGI-3}
\label{app:arc_full_results}

We report the complete evaluation results of MAP and ReAct (Claude 4.6) 
across the remaining game environments in ARC-AGI-3 in 
Table~\ref{tab:arc_full_results}. MAP achieves consistent improvements 
over ReAct in 22 out of 25 games in total, with ReAct scoring near-zero 
across virtually all environments. The subset of results reported in the 
main paper (Table~\ref{tab:arc_results}) was selected to provide a 
representative overview of performance across diverse game types; the 
full results here confirm that the improvements observed are broad and 
consistent rather than limited to a specific subset of games.

\begin{table}[h]
\centering
\small
\renewcommand{\arraystretch}{0.9}
\setlength{\tabcolsep}{3pt}           
\caption{Full results on the remaining 19 ARC-AGI-3 game environments.}
\vspace{-0.2cm}
\resizebox{0.60\textwidth}{!}{        
\begin{tabular}{l cc cc | l cc cc}
\toprule
\textbf{Game}
& \multicolumn{2}{c}{\textbf{ReAct}}
& \multicolumn{2}{c|}{\textbf{MAP}}
& \textbf{Game}
& \multicolumn{2}{c}{\textbf{ReAct}}
& \multicolumn{2}{c}{\textbf{MAP}} \\
\cmidrule(lr){2-3} \cmidrule(lr){4-5}
\cmidrule(lr){7-8} \cmidrule(lr){9-10}
& Lv. & Sc. & Lv. & Sc.
& & Lv. & Sc. & Lv. & Sc. \\
\midrule
BP35 & 0 & 0.00 & \cellcolor{blue!6}\textbf{0} & \cellcolor{blue!6}\textbf{0.00}
& CD82 & 0 & 0.00 & \cellcolor{blue!6}\textbf{2} & \cellcolor{blue!6}\textbf{3.08} \\
CN04 & 0 & 0.00 & \cellcolor{blue!6}\textbf{1} & \cellcolor{blue!6}\textbf{4.89}
& DC22 & 0 & 0.00 & \cellcolor{blue!6}\textbf{2} & \cellcolor{blue!6}\textbf{3.95} \\
FT09 & 0 & 0.00 & \cellcolor{blue!6}\textbf{2} & \cellcolor{blue!6}\textbf{3.98}
& G50T & 0 & 0.00 & \cellcolor{blue!6}\textbf{1} & \cellcolor{blue!6}\textbf{3.57} \\
KA59 & 1 & 3.57 & \cellcolor{blue!6}\textbf{1} & \cellcolor{blue!6}\textbf{4.75}
& LF52 & 2 & 5.09 & \cellcolor{blue!6}\textbf{2} & \cellcolor{blue!6}\textbf{6.19} \\
LP85 & 1 & 2.17 & \cellcolor{blue!6}\textbf{2} & \cellcolor{blue!6}\textbf{2.48}
& LS20 & 0 & 0.00 & \cellcolor{blue!6}\textbf{2} & \cellcolor{blue!6}\textbf{3.50} \\
M0R0 & 0 & 0.00 & \cellcolor{blue!6}\textbf{2} & \cellcolor{blue!6}\textbf{4.77}
& R11L & 0 & 0.00 & \cellcolor{blue!6}\textbf{1} & \cellcolor{blue!6}\textbf{3.27} \\
S5I5 & 0 & 0.00 & \cellcolor{blue!6}\textbf{2} & \cellcolor{blue!6}\textbf{7.83}
& SK48 & 0 & 0.00 & \cellcolor{blue!6}\textbf{1} & \cellcolor{blue!6}\textbf{2.05} \\
SP80 & 1 & 4.76 & \cellcolor{blue!6}\textbf{2} & \cellcolor{blue!6}\textbf{5.10}
& SU15 & 0 & 0.00 & \cellcolor{blue!6}\textbf{1} & \cellcolor{blue!6}\textbf{3.22} \\
TN36 & 1 & 3.57 & \cellcolor{blue!6}\textbf{1} & \cellcolor{blue!6}\textbf{4.57}
& TR87 & 0 & 0.00 & \cellcolor{blue!6}\textbf{0} & \cellcolor{blue!6}\textbf{0.00} \\
SC25 & 0 & 0.00 & \cellcolor{blue!6}\textbf{0} & \cellcolor{blue!6}\textbf{0.00}
& & & & & \\
\bottomrule
\end{tabular}
}
\label{tab:arc_full_results}
\end{table}

\subsection{Map QA Accuracy}
\label{app:map_qa}

To directly measure whether the mapping stage produces accurate and 
causally grounded environmental knowledge, we design an offline QA 
evaluation based on the constructed cognitive map $M_t$. The evaluation 
covers four categories of environment-probing questions:
\begin{itemize}[leftmargin=*]
    \item \textbf{Object Location}: queries about the spatial position 
    of a specific object (e.g., ``Where is the apple?''), evaluating 
    whether the agent has correctly mapped the spatial layout of the 
    environment.
    
    \item \textbf{Object-Action Affordance}: queries about the effect 
    of a given action on an object (e.g., ``What happens when you open 
    the fridge?''), evaluating whether the agent has correctly captured 
    action consequences during exploration.
    
    \item \textbf{Negative Knowledge}: queries about the absence of 
    objects in explored locations (e.g., ``Is there a knife in the 
    bedroom?''), evaluating whether the agent can correctly report 
    non-existence based on its map.
    
    \item \textbf{Task Reasoning}: queries about task-relevant 
    decisions derived from the map (e.g., ``Which receptacle should 
    the agent visit first to complete the task?''), evaluating whether 
    the map supports downstream planning.
\end{itemize}
For each question category, ground-truth answers are extracted directly 
from the environment engine's internal state. The agent is queried based 
solely on its constructed $M_t$, without access to additional interaction. 
Accuracy is computed as the proportion of correctly answered questions 
within each category, averaged across all evaluated task instances.

\subsection{Causal Adaptability under Environment Shift}
\label{app:perturbation}

To evaluate whether mapping-based training cultivates genuine causal 
understanding rather than surface-level pattern matching, we introduce 
controlled mid-episode perturbations: target objects are relocated to 
alternative positions at a randomly sampled step during task execution, 
requiring the agent to detect and adapt to the environmental change.

We design three metrics to characterize agent behavior under perturbation:
\begin{itemize}[leftmargin=*]
    \item \textbf{pass@1$_{\text{perturb}}$}: Task success rate under 
    perturbation conditions, computed identically to pass@1. A smaller 
    degradation from pass@1 to pass@1$_{\text{perturb}}$ indicates 
    greater robustness to environmental shifts.

    \item \textbf{Re-exploration Rate}: The proportion of perturbed 
    rollouts in which the agent executes at least one \texttt{go to} 
    action navigating to an alternative location after the perturbation 
    is triggered. This metric reflects whether the agent actively 
    re-explores the environment upon detecting displacement, rather than 
    persisting with invalid actions at the original location.

    \item \textbf{$\Delta$Steps}: The average number of interaction 
    steps from the perturbation point to episode termination (either 
    task success or step budget exhaustion). Formally, for each 
    perturbed rollout:
    \begin{equation}
        \Delta\text{Steps} = \frac{1}{|\mathcal{R}|} 
        \sum_{r \in \mathcal{R}} 
        \left( L_r - t^r_{\text{perturb}} - 1 \right),
    \end{equation}
    where $L_r$ is the total length of rollout $r$, 
    $t^r_{\text{perturb}}$ is the step index at which the perturbation 
    is triggered, and $\mathcal{R}$ denotes the set of all perturbed 
    rollouts. A smaller $\Delta$Steps indicates that the agent recovers 
    more efficiently after the perturbation, reflecting lower adaptation 
    cost.
\end{itemize}
The three metrics form a causal chain characterizing the 
agent's full recovery process: \textbf{pass@1$_{\text{perturb}}$} 
measures overall robustness, \textbf{Re-exploration Rate} captures 
whether the agent actively seeks the displaced object, and 
\textbf{$\Delta$Steps} quantifies the efficiency of recovery once 
re-exploration is initiated.

\section{Global Exploration Prompt}
\subsection{Focus Points Generation}
\label{app:prompt_global}

The Focus Points Generation prompt is used in Stage 1 to guide the 
agent in analyzing the task environment and deriving actionable 
exploration priorities prior to interaction.

\begin{tcolorbox}[colframe=black, 
    title=Focus Points Generation Prompt,
    colback=gray!5, fonttitle=\bfseries, breakable]
You are a Task Learning Guidance Agent whose mission is to analyze a 
given task environment and its examples in detail. Your objective is 
to derive a comprehensive list of actionable focus points that will 
guide subsequent exploration toward discovering environment-level 
general rules shared across all tasks.

When producing your analysis, please adhere to the following 
requirements:
\begin{itemize}[leftmargin=*]
    \item Your focus points must target the underlying interaction 
    logic of the environment, independent of specific task goals.
    \item Each focus point should provide clear guidance on what 
    interaction patterns, action syntax, constraints, error patterns, 
    and naming conventions to investigate during exploration.
    \item The analysis should help identify environment-level general 
    rules that are shared across all tasks, constituting the 
    foundational logic for interacting with this environment.
\end{itemize}

For each focus point you extract, follow the structure below:

First, explain your reasoning process that led you to identify this 
focus point. This reasoning should be based on a careful analysis of 
the provided environment description and example trajectories, 
focusing on recurring interaction patterns, failure modes, and 
environment-specific conventions.

Next, present the concrete focus point as a moderately detailed 
sentence that provides clear guidance for discovering environment-
general rules. The focus point must be specific and closely linked 
to the environment's interaction logic, but do not use ``such as'' 
or ``e.g.'' to explain it.

Finally, ensure that each pair of reasoning and focus point is 
numbered sequentially. For example:

\texttt{Reasoning 1: [Detailed explanation...]}\\
\texttt{Focus Point 1: [Specific guidance sentence...]}

\end{tcolorbox}

\subsection{Knowledge Distillation Prompt}
\label{app:prompt_distill}

The Knowledge Distillation prompt is used at the end of Stage 1 to 
distill accumulated exploration trajectories into structured 
environment-general rules $K_g$, capturing the underlying interaction 
logic shared across all tasks.

\begin{tcolorbox}[colframe=black, 
    title=Knowledge Distillation Prompt,
    colback=gray!5, fonttitle=\bfseries, breakable]
You are an Environment Knowledge Distillation Agent. By examining both 
successful and failed exploration trajectories of the same environment, 
your goal is to discover environment-level general rules that are shared 
across all tasks. These rules should be independent of specific task 
goals and capture the underlying interaction logic of the environment.

When analyzing the trajectories, focus on the following three dimensions:
\begin{itemize}[leftmargin=*]
    \item \textbf{Action Syntax}: Identify the correct format and 
    structure of valid actions in this environment, including how 
    actions should be phrased and what arguments they require.
    \item \textbf{Interaction Rules}: Identify the preconditions and 
    constraints that govern valid action execution, including 
    sequencing requirements, capacity limits, and physical or logical 
    dependencies between actions.
    \item \textbf{Error Patterns}: Identify recurring failure modes 
    observed across trajectories, including common invalid actions, 
    inapplicable commands, and the environmental constraints that 
    caused them.
\end{itemize}

The rules you generate must be:
\begin{itemize}[leftmargin=*]
    \item \textbf{Environment-general}: applicable across different 
    task instances in the same environment, not tied to any specific 
    task goal.
    \item \textbf{Grounded}: directly supported by evidence from the 
    provided trajectories. Avoid speculative or unverified statements.
    \item \textbf{Non-redundant}: ensure no two rules overlap in 
    meaning or scope.
    \item \textbf{Concise}: a maximum of 15 rules. Prioritize the most 
    impactful and generalizable patterns.
\end{itemize}
\end{tcolorbox}

\section{Task Mapping Prompt}
\label{app:prompt_mapping}

Stage 2 employs a Role-Purpose-Priority protocol to guide 
the agent in constructing a task-specific cognitive map $M_t$ prior to 
execution. The Alfworld, TextCraft and ScienceWorld versions are provided below.

\begin{figure}[H]
    \centering
    \small
    \begin{tcolorbox}[colback=gray!5!white, colframe=black!15, width=\linewidth, 
    boxrule=0.5pt, arc=2mm]
    
    \textbf{System:} You are a \colorEvidence{Task-oriented Scout}. Your mission is \colorEvidence{NOT} to complete 
    the task, but to systematically explore the environment and construct 
    a structured cognitive map for a downstream executor. \\
    \textbf{Exploration Priorities:} \\
    \ding{182} Map the environmental distribution: identify rooms, 
    locations, and objects present in the environment, along with their 
    spatial relationships and reachability; \\
    \ding{183} Probe action consequences: interact with key objects to 
    observe how the environment responds; \\
    \ding{184} Record task-relevant affordances: document object states, 
    container contents, and interaction preconditions that are critical 
    for downstream task execution. \\
    \textbf{Constraints:} \\
    \ding{182} \colorEvidence{Gather information rather than complete the task}; \\
    \ding{183} Upon failure, diagnose the cause and attempt alternative 
    actions or locations; \\
    \ding{184} \colorEvidence{Ground your exploration in the global knowledge 
    $K_g$ to avoid redundant interactions.}

    \end{tcolorbox}
    \caption{Task mapping prompt skeleton for ALFWorld (Stage 2).}
    \label{fig:task_mapping_prompt}
\end{figure}

\begin{figure}[H]
    \centering
    \small
    \begin{tcolorbox}[colback=gray!5!white, colframe=black!15, width=\linewidth, 
    boxrule=0.5pt, arc=2mm]
    
    \textbf{System:} You are a \colorEvidence{Task-oriented Scout}. 
    Your mission is \colorEvidence{NOT} to craft the target item, but 
    to systematically explore the crafting environment and construct 
    a structured cognitive map for a downstream executor. \\
    \textbf{Exploration Priorities:} \\
    \ding{182} Map the spatial layout of the crafting environment: 
    identify all accessible crafting stations, storage locations, and 
    resource nodes, along with their connectivity and reachability; \\
    \ding{183} Catalog object-action affordances: interact with crafting 
    stations and ingredient sources to observe what actions are 
    applicable and what state changes each action produces; \\
    \ding{184} Record task-relevant affordance preconditions: document 
    ingredient availability, quantity requirements, and crafting station 
    conditions critical for downstream task execution. \\
    \textbf{Constraints:} \\
    \ding{182} \colorEvidence{Gather spatial and affordance information 
    rather than complete the target recipe}; \\
    \ding{183} Upon failure, diagnose the cause and attempt alternative 
    actions or locations; \\
    \ding{184} \colorEvidence{Ground your exploration in the global 
    knowledge $K_g$ to avoid redundant interactions.}

    \end{tcolorbox}
    \caption{Task mapping prompt skeleton for TextCraft (Stage 2).}
    \label{fig:task_mapping_prompt_textcraft}
\end{figure}

\begin{figure}[H]
    \centering
    \small
    \begin{tcolorbox}[colback=gray!5!white, colframe=black!15, width=\linewidth, 
    boxrule=0.5pt, arc=2mm]
    
    \textbf{System:} You are a \colorEvidence{Task-oriented Scout}. 
    Your mission is \colorEvidence{NOT} to complete the scientific 
    experiment, but to systematically explore the laboratory environment 
    and construct a structured cognitive map for a downstream executor. \\
    \textbf{Exploration Priorities:} \\
    \ding{182} Map the spatial layout of the laboratory: identify all 
    accessible locations, their 
    topological connectivity, and the objects and instruments present 
    at each location; \\
    \ding{183} Catalog object-action affordances: interact with 
    instruments, substances, and containers to observe what actions are 
    applicable and what physical or chemical state changes each action 
    produces; \\
    \ding{184} Record task-relevant affordance preconditions: document 
    object properties, 
    container contents, and instrument conditions critical for downstream 
    experimental execution. \\
    \textbf{Constraints:} \\
    \ding{182} \colorEvidence{Gather spatial and affordance information 
    rather than complete the experiment}; \\
    \ding{183} Upon failure, diagnose the cause and attempt alternative 
    actions or locations; \\
    \ding{184} \colorEvidence{Ground your exploration in the global 
    knowledge $K_g$ to avoid redundant interactions.}

    \end{tcolorbox}
    \caption{Task mapping prompt skeleton for ScienceWorld (Stage 2).}
    \label{fig:task_mapping_prompt_sciworld}
\end{figure}

\section{Knowledge-Augmented Execution Prompt}
\label{app:prompt_acting}

Stage 3 injects both the global knowledge $K_g$ and the task-specific 
cognitive map $M_t$ into the acting prompt as contextual priors. 
$K_g$ provides environment-general interaction rules shared across all 
task instances, while $M_t$ supplies task-specific spatial layouts and 
object-action affordances constructed during the mapping stage. Together, 
they enable the agent to perform knowledge-driven execution without 
additional exploration. The execution prompts for all three environments 
are provided below.

\begin{tcolorbox}[colframe=black,
    title=Knowledge-Augmented Execution Prompt for ALFWorld (Stage 3),
    colback=gray!5, fonttitle=\bfseries, breakable]

\textbf{System:} You are a helpful assistant controlling a robot in a 
household environment. Your task is to complete household tasks by 
navigating rooms and interacting with objects.\\
\textbf{Available Actions:}\\
\textit{Navigation \& Observation:} \texttt{look}, \texttt{inventory}, 
\texttt{go to (receptacle)}\\
\textit{Receptacle Interaction:} \texttt{open/close (receptacle)}\\
\textit{Object Interaction:} \texttt{take (object) from (receptacle)}, 
\texttt{move (object) to (receptacle)}, \texttt{examine (object)}\\
\textit{State Changes:} \texttt{heat/cool/clean (object) with 
(receptacle)}, \texttt{slice (object) with (object)}\\
\textbf{Global Knowledge} [$K_g$\textbf{]:}\\
\texttt{\{global\_knowledge\}}\\
\textbf{Cognitive Map} [$M_t$\textbf{]:}\\
\texttt{\{mental\_map\}}\\
\textbf{Important Rules:}
\begin{itemize}[leftmargin=*, nosep]
    \item Navigate to a location before interacting with objects there;
    \item You can only hold one object at a time;
    \item Always check admissible commands to see what actions are 
    currently valid.
\end{itemize}
\end{tcolorbox}

\vspace{1em}

\begin{tcolorbox}[colframe=black,
    title=Knowledge-Augmented Execution Prompt for TextCraft (Stage 3),
    colback=gray!5, fonttitle=\bfseries, breakable]

\textbf{System:} You are an expert Minecraft crafter. Your task is to 
craft a specific target item using the available crafting recipes.\\
\textbf{Available Actions:}\\
\texttt{get [quantity] [item]}: Gather raw materials that cannot be 
crafted\\
\texttt{craft [quantity] [item] using [ingredients]}: Craft an item 
using ingredients from your inventory\\
\texttt{inventory}: Check what items you currently have\\
\textbf{Global Knowledge} [$K_g$\textbf{]:}\\
\texttt{\{global\_knowledge\}}\\
\textbf{Cognitive Map} [$M_t$\textbf{]:}\\
\texttt{\{mental\_map\}}\\
\textbf{Important Rules:}
\begin{itemize}[leftmargin=*, nosep]
    \item You can only \texttt{get} raw materials that cannot be 
    crafted from other items;
    \item Craft intermediate items first before crafting the final 
    target;
    \item Item names use spaces instead of underscores 
    (e.g., \texttt{oak planks} not \texttt{oak\_planks}).
\end{itemize}
\end{tcolorbox}

\vspace{1em}

\begin{tcolorbox}[colframe=black,
    title=Knowledge-Augmented Execution Prompt for ScienceWorld (Stage 3),
    colback=gray!5, fonttitle=\bfseries, breakable]

\textbf{System:} You are a scientist performing experiments in a 
virtual science laboratory. Your task is to complete the assigned 
experiment by interacting with objects in the environment.\\
\textbf{Available Actions:}\\
\textit{Observation:} \texttt{look around}, \texttt{look at [object]}, 
\texttt{look in [container]}, \texttt{inventory}\\
\textit{Object Interaction:} \texttt{pick up/put down [object]}, 
\texttt{put [object] in [container]}, \texttt{open/close [object]}\\
\textit{Device Operation:} \texttt{activate/deactivate [device]}, 
\texttt{use [object] on [object]}, \texttt{connect/disconnect 
[object] to [object]}\\
\textit{Substance Handling:} \texttt{pour [substance] in [container]}, 
\texttt{mix [container]}, \texttt{dunk [object] in [container]}\\
\textit{Navigation:} \texttt{teleport to [room]}\\
\textbf{Global Knowledge} [$K_g$\textbf{]:}\\
\texttt{\{global\_knowledge\}}\\
\textbf{Cognitive Map} [$M_t$\textbf{]:}\\
\texttt{\{mental\_map\}}\\
\textbf{Important Rules:}
\begin{itemize}[leftmargin=*, nosep]
    \item Navigate to the correct location before interacting with 
    objects there;
    \item Some experiments require specific action sequences 
    (e.g., heat water, then measure temperature);
    \item Use \texttt{look around} to observe surroundings when 
    entering a new location.
\end{itemize}
\end{tcolorbox}

\section{CoMAP Baseline Prompt}
\label{app:comap_prompt}

CoMAP (Map-and-Act) is a non-staged variant that shares the same base
system prompt as ReAct, but additionally instructs the agent to
\textbf{simultaneously} maintain an internal world model and execute
the task within a single interaction loop---without a dedicated mapping
stage. We provide the ALFWorld version as a representative example;
TextCraft and ScienceWorld follow an identical structure with
environment-specific action spaces.

\begin{tcolorbox}[colframe=black,
    title=CoMAP Prompt (ALFWorld),
    colback=gray!5, fonttitle=\bfseries, breakable]
\textbf{System:} You are a helpful assistant controlling a robot in a
household environment. Your task is to complete household tasks by
navigating rooms and interacting with objects.\\
\textbf{Available Actions:}\\
\textit{Navigation \& Observation:}
\texttt{look}, \texttt{inventory}, \texttt{go to (receptacle)}\\
\textit{Receptacle Interaction:}
\texttt{open/close (receptacle)}\\
\textit{Object Interaction:}
\texttt{take (object) from (receptacle)},
\texttt{move (object) to (receptacle)},
\texttt{examine (object)}\\
\textit{State Changes:}
\texttt{heat/cool/clean (object) with (receptacle)},
\texttt{slice (object) with (object)}\\[0.5em]
\textbf{Map-and-Act Strategy}\\
As you execute the task, you should \textbf{simultaneously} build and
maintain an internal world model of the environment. Specifically:
\begin{itemize}[leftmargin=*, nosep]
    \item \textbf{Track spatial layout:} Record which receptacles and
    objects you have observed, and update your mental map as you
    navigate.
    \item \textbf{Log object-action affordances:} Note which actions
    are valid for each object or receptacle encountered, and remember
    failed attempts to avoid repetition.
    \item \textbf{Reason from your world model:} Before each action,
    consult your accumulated environmental knowledge to make informed
    decisions rather than acting blindly.
    \item \textbf{Balance mapping and execution:} Do not explore
    exhaustively before acting---instead, interleave targeted
    exploration with task progress, updating your world model
    incrementally.
\end{itemize}
\textbf{Important Rules:}
\begin{itemize}[leftmargin=*, nosep]
    \item Navigate to a location before interacting with objects there;
    \item You can only hold one object at a time;
    \item Always check admissible commands to see what actions are
    currently valid.
\end{itemize}
\end{tcolorbox}

\noindent The key distinction from MAP is that CoMAP does not separate
the mapping and execution phases---the agent is required to
simultaneously gather environmental knowledge and complete the task
within a single interaction loop. This design directly contrasts with
MAP's staged approach, where a dedicated mapping phase constructs an
explicit cognitive map $M_t$ prior to task execution, enabling more
structured and informed decision-making.

\section{Case Study}
\label{app:case_study}

\subsection{Long-Horizon Interactive Benchmarks}
\label{app:case_structured}

We present representative trajectories from ALFWorld, TextCraft, and 
ScienceWorld to illustrate how the two-stage knowledge injection 
($K_g$ and $M_t$) enables efficient task execution.
\begin{figure}[H]
\centering
\includegraphics[width=\linewidth]{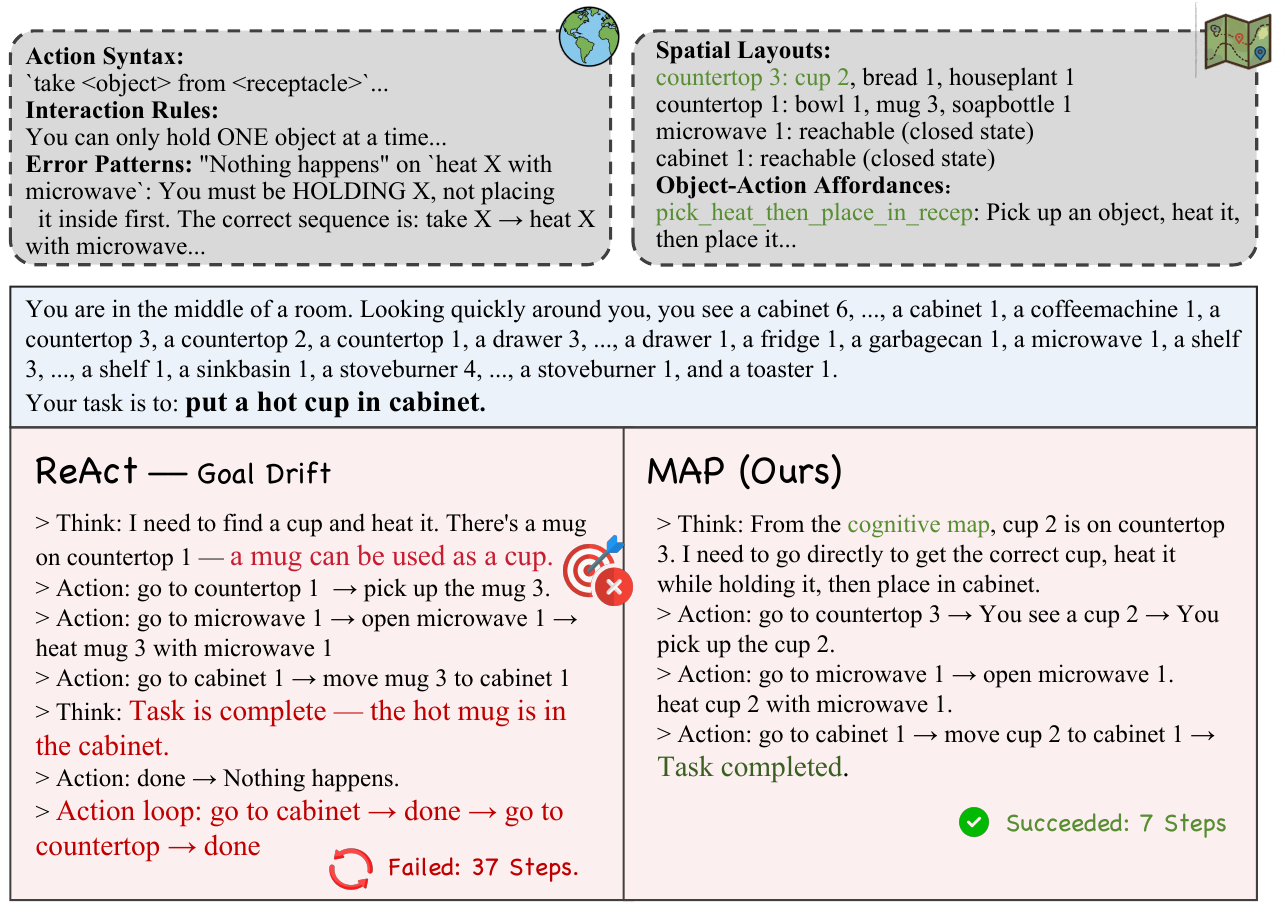}
\caption{\textbf{Case Study on ALFWorld.} ReAct misidentifies \texttt{mug} as the target \texttt{cup} due to absent environmental priors, 
falling into a futile action loop and failing after \textbf{37 steps}. 
MAP leverages its pre-constructed cognitive map to precisely locate \texttt{cup 2}, 
completing the task in just \textbf{7 steps} ($\uparrow$81\% efficiency), 
validating the core value of the ``Let's look around first'' paradigm.} 
\label{fig:case_study_1}
\end{figure}

\begin{figure}[H]
\centering
\includegraphics[width=\linewidth]{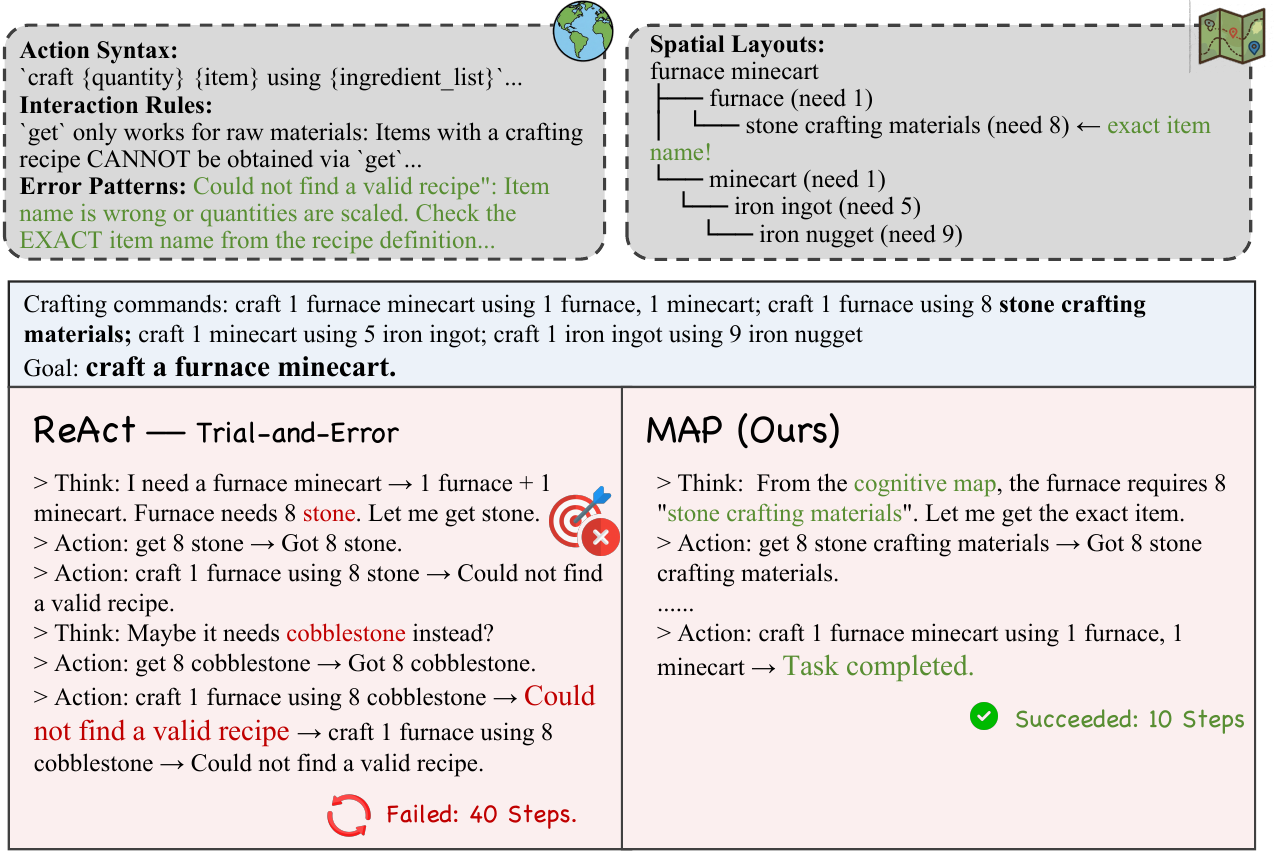}
\caption{\textbf{Case Study on Textcraft.} ReAct repeatedly guesses incorrect ingredient names (\texttt{stone}, \texttt{cobblestone}) 
due to absent environmental priors, falling into a trial-and-error loop and failing after \textbf{40 steps}. 
MAP precisely identifies the correct ingredient \texttt{stone crafting materials} 
from its pre-constructed cognitive map, completing the task in just \textbf{10 steps} ($\uparrow$75\% efficiency).} 
\label{fig:case_study_2}
\end{figure}

\begin{figure}[H]
\centering
\includegraphics[width=\linewidth]{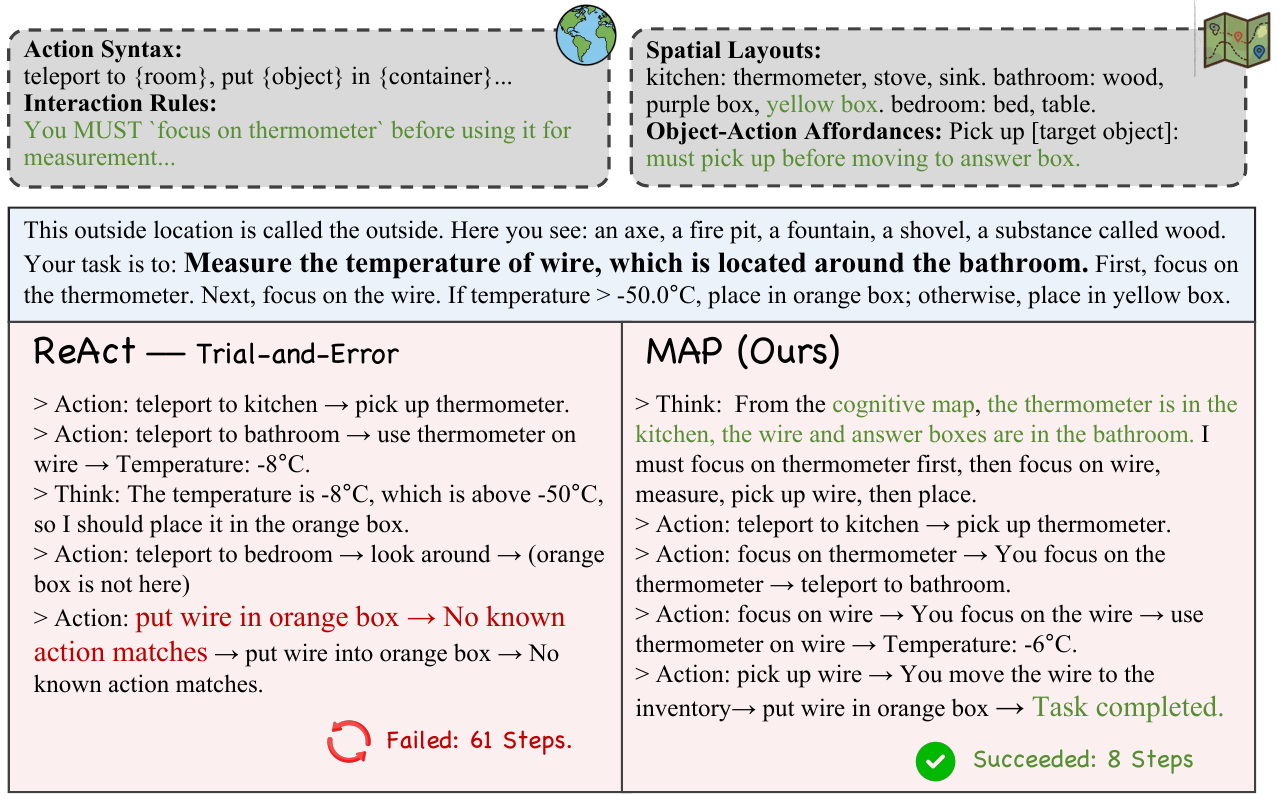}
\caption{\textbf{Case Study on Scienceworld.} ReAct fails to locate the target container and repeatedly triggers invalid actions 
due to unknown spatial layouts and interaction constraints, failing after \textbf{61 steps}. 
MAP directly retrieves object locations and prerequisite action sequences 
from its pre-constructed cognitive map, completing the task in just \textbf{8 steps} ($\uparrow$87\% efficiency).} 
\label{fig:case_study_3}
\end{figure}

\subsection{Fluid Intelligence Benchmarks}
\label{app:case_arcagi3}

We present representative cognitive maps $M_t$ constructed by MAP 
on three ARC-AGI-3 game instances (Table~\ref{tab:arc_cognitive_maps}), 
covering games of fundamentally different mechanics: maze navigation 
(TU93), belt alignment (VC33), and color sorting (SB26). Across all 
three instances, MAP autonomously discovers spatial layouts, action 
effects, and game rules through structured exploration---directly 
enabling the agent to formulate winning strategies and advance through 
multiple levels without any explicit instructions or goals.

\begin{table*}[t]
\centering
\caption{\textbf{Representative Cognitive Maps $M_t$ Constructed by 
MAP on ARC-AGI-3.}}
\label{tab:arc_cognitive_maps}
\resizebox{\textwidth}{!}{%
\begin{tabular}{p{2.8cm} p{5.2cm} p{4.8cm} p{5.2cm}}
\toprule
\textbf{Game} 
& \textbf{Environment Layout} 
& \textbf{Action Effects} 
& \textbf{Game Rules} \\
\midrule

\rowcolor{gray!10}
\multicolumn{4}{l}{\texttt{Grid Maze Navigation}} \\

\textit{TU93}
& 6$\times$6 node maze on 64$\times$64 grid; player (color 9) 
starts at (0,0); target (color 14) at (5,5); color 2 = corridors, 
color 5 = walls; bottom row = move counter (color 6, 
right-to-left).
& ACTION1--4: move up/down/left/right by 6px; requires color 2 
corridor, else blocked; straight move costs 1 counter pixel; 
turning costs 2; blocked moves still consume counter.
& Reach color 14 to win; counter exhaustion = failure; hazard 
blocks (color 8/12) move one step per successful player move, 
bounce at dead ends (period $= 2\times(\text{path}-1)$); 
player-hazard collision = GAME\_OVER. \\

\midrule
\rowcolor{gray!10}
\multicolumn{4}{l}{\texttt{Scrolling Belt Alignment Puzzle}} \\

\textit{VC33}
& Upper belt (rows 1--19) and lower belt (rows 24+) separated by 
fixed walls (color 5); four buttons (color 9) on left edge control 
belt scrolling; lower belt contains movable objects (color 4 + 14); 
wall gaps (color 14) = target positions; top row = counter 
(color 7).
& ACTION6 + button coordinates scrolls belt 4px: (2,18) = 
upper-right, (2,26) = upper-left, (2,38) = lower-left, (2,46) = 
lower-right; coupled buttons move two objects simultaneously; 
non-button clicks waste counter.
& Belts scroll independently; fixed walls immovable; win = align 
object with wall gap; floor-as-fuel mechanic: movement consumes 
color 0 floor tiles, limiting range; buttons in UP/DOWN pairs, 
each pair controls one element. \\

\midrule
\rowcolor{gray!10}
\multicolumn{4}{l}{\texttt{Color Sorting Placement Puzzle}} \\

\textit{SB26}
& Top panel (rows 0--7): target color sequence; middle frame: 
empty slots (color 2), 4 slots at Level 0, 7 slots at Level 2 
(dual frame); bottom palette: N shuffled colored blocks 
(4$\times$4); row 53 = move counter (1 pixel per placement, 
color 2$\to$3).
& ACTION6 + palette block: select block, free; ACTION6 + empty 
slot: place selected color, costs 1 step; ACTION6 + filled slot: 
reselect for repositioning, free; ACTION5: submit solution, costs 
1 step; ACTION6 (no coords) / ACTION7: no effect.
& Match palette blocks to target sequence left-to-right; incorrect 
placements are repositionable; ACTION5 required to complete level; 
Level 0: 4-color single frame; Level 2: 7-color dual frame with 
interleaved slot mapping. \\

\bottomrule
\end{tabular}%
}
\end{table*}


\end{document}